\documentclass{article}

\usepackage{arxiv}

\usepackage[utf8]{inputenc} 
\usepackage[T1]{fontenc}    
\usepackage{hyperref}       
\usepackage{url}            
\usepackage{booktabs}       
\usepackage{amsfonts}       
\usepackage{nicefrac}       
\usepackage{microtype}      
\usepackage{lipsum}
\usepackage{graphicx}
\usepackage{indentfirst}
\usepackage{multirow}
\usepackage{siunitx}  
\usepackage{amssymb}  
\usepackage{amsmath}
\usepackage{makecell}

\usepackage[utf8]{inputenc}
\usepackage[T1]{fontenc}
\usepackage[english]{babel}
\usepackage{soul}

\usepackage{microtype}      

\usepackage[
    square,
    numbers,
    sort&compress
]{natbib}

\usepackage{tikz}
\usepackage[edges]{forest}

\definecolor{paired-light-blue}{RGB}{198, 219, 239}
\definecolor{paired-dark-blue}{RGB}{49, 130, 188}
\definecolor{paired-light-orange}{RGB}{251, 208, 162}
\definecolor{paired-dark-orange}{RGB}{230, 85, 12}
\definecolor{paired-light-green}{RGB}{199, 233, 193}
\definecolor{paired-lightcl-green}{RGB}{209, 243, 203}
\definecolor{paired-lightot-green}{RGB}{179, 213, 173}
\definecolor{paired-dark-green}{RGB}{49, 163, 83}
\definecolor{paired-light-purple}{RGB}{218, 218, 235}
\definecolor{paired-dark-purple}{RGB}{117, 107, 176}
\definecolor{paired-light-gray}{RGB}{217, 217, 217}
\definecolor{paired-dark-gray}{RGB}{99, 99, 99}
\definecolor{paired-light-pink}{RGB}{222, 158, 214}
\definecolor{paired-dark-pink}{RGB}{123, 65, 115}
\definecolor{paired-light-red}{RGB}{231, 150, 156}
\definecolor{paired-dark-red}{RGB}{131, 60, 56}
\definecolor{paired-light-yellow}{RGB}{231, 204, 149}
\definecolor{paired-dark-yellow}{RGB}{141, 109, 49}
\tikzset{%
    parent/.style = {align=center,text width=2cm,rounded corners=3pt, line width=0.3mm, fill=gray!10,draw=gray!80},
    top_class/.style = {align=center,text width=3cm,rounded corners=3pt, fill=paired-light-gray!50,draw=paired-dark-gray!65,line width=0.3mm},
    rnn/.style = {align=center,text width=3cm,rounded corners=3pt, fill= paired-light-green!50,draw=paired-dark-green!75,line width=0.3mm},
    rnn_more/.style = {align=center,text width=4cm,rounded corners=3pt, fill= paired-light-green!50,draw=paired-dark-green!75,line width=0.3mm},   
    rnn_work/.style = {align=center,text width=4.5cm,rounded corners=3pt, fill= paired-light-green!50,draw= cyan!0,line width=0.3mm},
    rnn_more_cl/.style = {align=center,text width=4cm,rounded corners=3pt, fill= paired-lightcl-green!50,draw=paired-dark-green!75,line width=0.3mm},   
    rnn_work_cl/.style = {align=center,text width=4.5cm,rounded corners=3pt, fill= paired-lightcl-green!50,draw= cyan!0,line width=0.3mm},
    rnn_more_ot/.style = {align=center,text width=4cm,rounded corners=3pt, fill= paired-lightot-green!50,draw=paired-dark-green!75,line width=0.3mm},   
    rnn_work_ot/.style = {align=center,text width=4.5cm,rounded corners=3pt, fill= paired-lightot-green!50,draw= cyan!0,line width=0.3mm},
    gnn/.style = {align=center,text width=3cm,rounded corners=3pt, fill=paired-light-orange!50,draw=paired-dark-orange!65,line width=0.3mm},  
    gnn_more/.style = {align=center,text width=4cm,rounded corners=3pt, fill=paired-light-orange!50,draw=paired-dark-orange!65,line width=0.3mm}, 
    gnn_work/.style = {align=center,text width=4.5cm,rounded corners=3pt, fill=paired-light-orange!50,draw=red!0,line width=0.3mm},    
    transformer/.style = {align=center,text width=3cm,rounded corners=3pt, fill=paired-light-blue!50,draw=paired-dark-blue!65,line width=0.3mm},
    transformer_more/.style = {align=center,text width=4cm,rounded corners=3pt, fill=paired-light-blue!50,draw=paired-dark-blue!65,line width=0.3mm},   
    transformer_work/.style = {align=center, text width=4.5cm,rounded corners=3pt, fill=paired-light-blue!50,draw=blue!0,line width=0.3mm},
    cnn/.style = {align=center,text width=3cm,rounded corners=3pt, fill= paired-light-purple!50,draw=paired-dark-purple!75,line width=0.3mm},   
    cnn_more/.style = {align=center,text width=4cm,rounded corners=3pt, fill= paired-light-purple!50,draw=paired-dark-purple!75,line width=0.3mm},
    cnn_work/.style = {align=center,text width=4.5cm,rounded corners=3pt, fill= paired-light-purple!50,draw= orange!0,line width=0.3mm},       
    multi/.style = {align=center,text width=3cm,rounded corners=3pt, fill= paired-light-red!35,draw=paired-dark-red!90,line width=0.3mm},           
    multi_more/.style = {align=center,text width=4cm,rounded corners=3pt, fill= paired-light-red!35,draw=paired-dark-red!90,line width=0.3mm},
    multi_work/.style = {align=center,text width=4.5cm,rounded corners=3pt, fill= paired-light-red!35,draw= magenta!0,line width=0.3mm}   
}

\title{Impact of Domain Knowledge and Multi-Modality on Intelligent Molecular Property Prediction: A Systematic Survey}

\author{
  Taojie Kuang \\
  Peng Cheng Laboratory\\
  South China University of Technology\\
  \\
  \And
  Pengfei Liu \\
  Peng Cheng Laboratory\\
  Sun Yat-Sen University\\
  \\
  \And
  Zhixiang Ren\thanks{Corresponding author} \\
  Peng Cheng Laboratory\\
  \texttt{renzhx@pcl.ac.cn} \\
}

\begin{document}
\maketitle

\begin{abstract}
The precise prediction of molecular properties is essential for advancements in drug development, particularly in virtual screening and compound optimization. 
The recent introduction of numerous deep learning-based methods has shown remarkable potential in enhancing molecular property prediction (MPP), especially improving accuracy and insights into molecular structures. 
Yet, two critical questions arise: does the integration of domain knowledge augment the accuracy of molecular property prediction and does employing multi-modal data fusion yield more precise results than unique data source methods? 
To explore these matters, we comprehensively review and quantitatively analyze recent deep learning methods based on various benchmarks.
We discover that integrating molecular information significantly improves molecular property prediction (MPP) for both regression and classification tasks. Specifically, regression improvements, measured by reductions in root mean square error (RMSE), are up to 4.0\%, while classification enhancements, measured by the area under the receiver operating characteristic curve (ROC-AUC), are up to 1.7\%. 
We also discover that enriching 2D graphs with 1D SMILES boosts multi-modal learning performance for regression tasks by up to 9.1\%, and augmenting 2D graphs with 3D information increases performance for classification tasks by up to 13.2\%, with both enhancements measured using ROC-AUC.
The two consolidated insights offer crucial guidance for future advancements in drug discovery.
\end{abstract}
\section{Introduction}
\label{s:introduction}
\noindent
The field of drug development has always been at the forefront of adopting innovative scientific techniques to enhance the discovery and optimization of therapeutic compounds. 
Central to this process is the prediction of molecular properties, a task that bears significant implications for drug screening and compound optimization\cite{shen2019molecular}.
Accurately predicting key molecular properties can significantly reduce the time and resources required in drug development, thereby hastening the journey towards innovative medical treatments.\\
\indent In the landscape of computational methods for molecular property prediction (MPP), deep learning (DL) has recently emerged as a transformative force, distinguishing itself markedly from traditional techniques such as quantitative structure-activity relationships (QSAR) and molecular dynamics simulations.
While conventional methods have laid the groundwork, DL significantly advances accuracy and analysis depth, enabling a more intricate exploration of the relationships between molecular structures and their properties\cite{li2022deep}.\\
\begin{figure*}[t]
    \centering
    \includegraphics[width=1.0\linewidth]{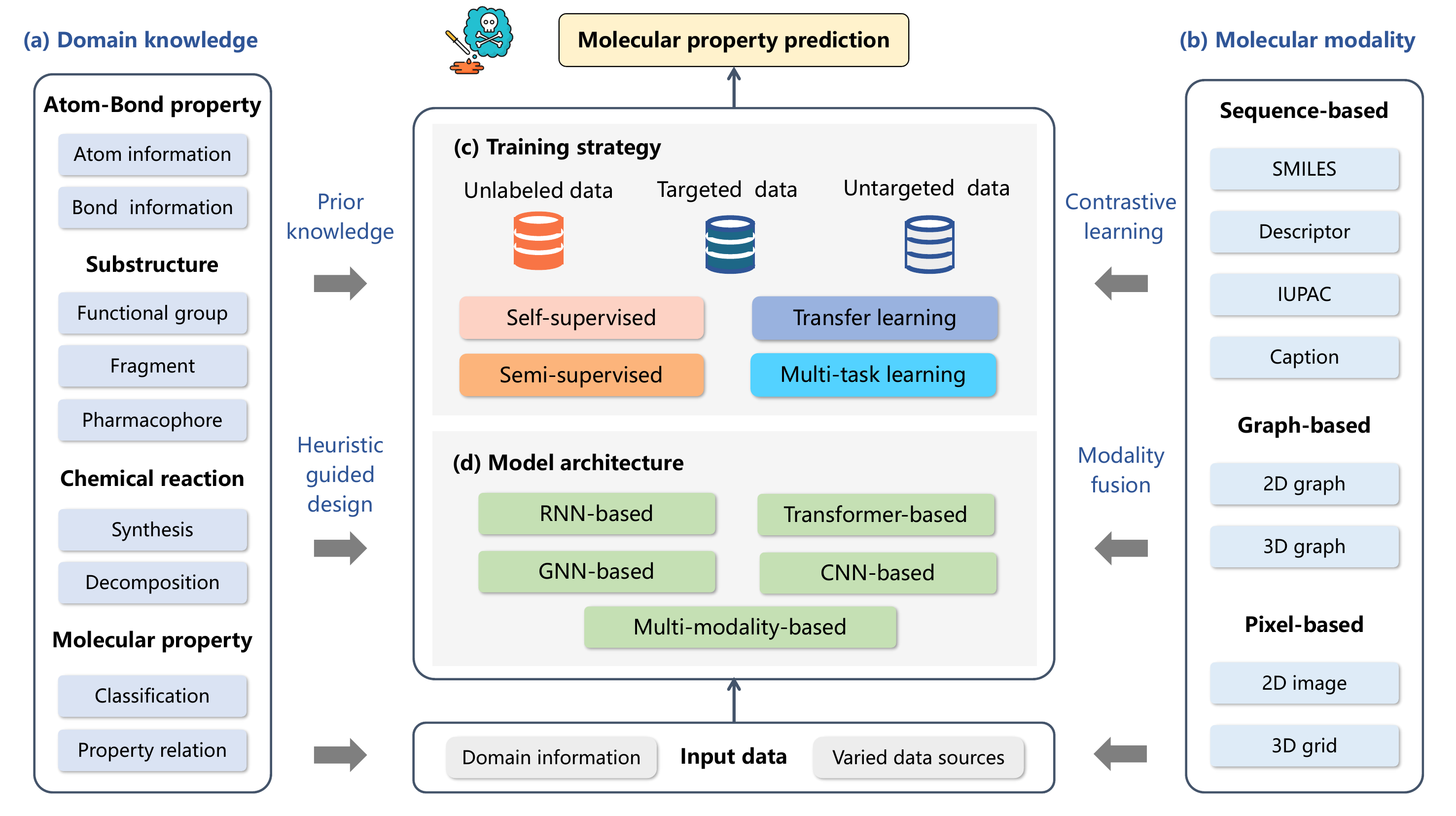}
    \caption{
            \textbf{The overview of our survey.} 
            we review the impact of domain knowledge and multi-modality on molecular property prediction from three critical aspects: input data, model architectures, and training strategy. 
            The detailed information are explained in the following sections.}
    \label{fig:OV}
\end{figure*}
\indent Despite the advancements in DL for MPP, the field continues to face ongoing evolution and challenges. Two significant trends are currently shaping the field. 
The first trend is the increasing integration of domain knowledge into DL models. 
This includes a broad spectrum of scientific information, such as chemical and physical property relation, atom and bond characteristics, and detailed insights into functional groups and molecular fragments. 
The integration of this knowledge aims to enhance the predictive accuracy of these models. 
This leads us to the critical question: Does more comprehensive domain knowledge actually improve the effectiveness of MPP? 
The second trend is the rising adoption of multi-modality techniques, which involve the fusion of various data types like sequence-based, graph-based and pixel-based formats. 
This approach is driven by the goal of achieving more accurate predictions in a field characterized by its complexity and data diversity, prompting the question: Is multi-modality more effective for MPP than methodologies that rely on uni-model data source? 
To explore these questions, our paper begins with an in-depth review of the current DL approaches in MPP, focusing on how the domain knowledge and multi-modal data integrate on encoder architecture and training strategy.\\
\indent Our review begins with an examination of various unique data encoder architectures for MPP such as Recurrent Neural Network (RNN)\cite{ma2020improving, lin2020novel, lv2021mol2context}, Graph Neural Network (GNN)\cite{han2023himgnn, bouritsas2022improving, song2020communicative, li2022kpgt, xiong2019pushing}, Transformer\cite{rong2020self, ross2022large, yin2023lgi, luo2022one}, and Convolutional Neural Network (CNN)\cite{zeng2022accurate, chen2021different, liu2019multiresolution} models, and also review the multi-modal methods\cite{liu2021pre, li2022geomgcl, zhu2022unified, guo2020graseq}.
We focus on how these architectures are aligned with existing molecular structural knowledge and their integration of domain knowledge. 
This exploration highlights the synergy between advanced computational techniques and fundamental molecular understanding, a crucial aspect in enhancing the accuracy of MPP.
Also we review a variety of training strategies, such as self-supervised\cite{wang2022molecular, fang2023knowledge, li2023knowledge}, semi-supervised\cite{hao2020asgn, sun2019infograph, zhang2023dropconn}, transfer learning\cite{sun2022pemp, chen2022meta, zhuang2023graph} and multi-tasks learning\cite{biswas2023predicting, tan2021multitask}. 
A particular emphasis is placed on strategies that effectively utilize unlabeled data, a vital consideration given the frequent scarcity of labeled data in this domain, and we focus on how the domain knowledge and multi-modal data to be used in the training strategy. 
Accompanying this review are comprehensive diagrams that systematically elucidate the nuances of these encoder architectures and training strategies, offering a clearer understanding of their complex mechanisms.
The overview of our paper is as Figure \ref{fig:OV}.\\
\indent Our study then proceeds to empirically evaluate these DL methods, utilizing pivotal benchmarks like MoleculeNet\cite{wu2018moleculenet}. 
These benchmarks, encompassing a diverse range of datasets each focused on specific molecular properties, allow for an extensive assessment of different DL approaches. 
A key aspect of our analysis is determining the impact of multi-modality techniques versus single modeling. 
Specifically, we investigate the effectiveness of integrating atom-bond level domain knowledge and substructures, such as functional groups and fragments, into the models. 
Additionally, we quantify the contributions of different data formats and conduct experiments to ascertain whether multi-modal fusion can enhance the generalization performance of the models. 
This evaluation not only provides comparative insights into the varied methods but also seeks to pinpoint essential factors that bolster the efficacy of DL in MPP.\\
\indent In summary, our main contributions are as follows:\\
\indent $\bullet$ We identify two pivotal issues when applying DL for MPP: domain knowledge integration and multi-modal data utilization.\\
\indent $\bullet$ We comprehensively review DL methods for MPP, featuring in-depth analyses of encoder architectures and training strategies.\\
\indent $\bullet$ We discover that integrating molecular substructure information results in a 4.0\% improvement on average in regression tasks and a 1.7\% increase on average in classification tasks.\\
\indent $\bullet$ We discover that enriching 2D graph models with 1D SMILES or 3D information boosts multi-modal learning, enhancing performance by 9.1\% to 13.2\% over single-modality models.
\section{Molecular Modality}
\label{s:DataFormat}
\begin{figure*}[t]
    \centering
    \includegraphics[width=1.0\linewidth]{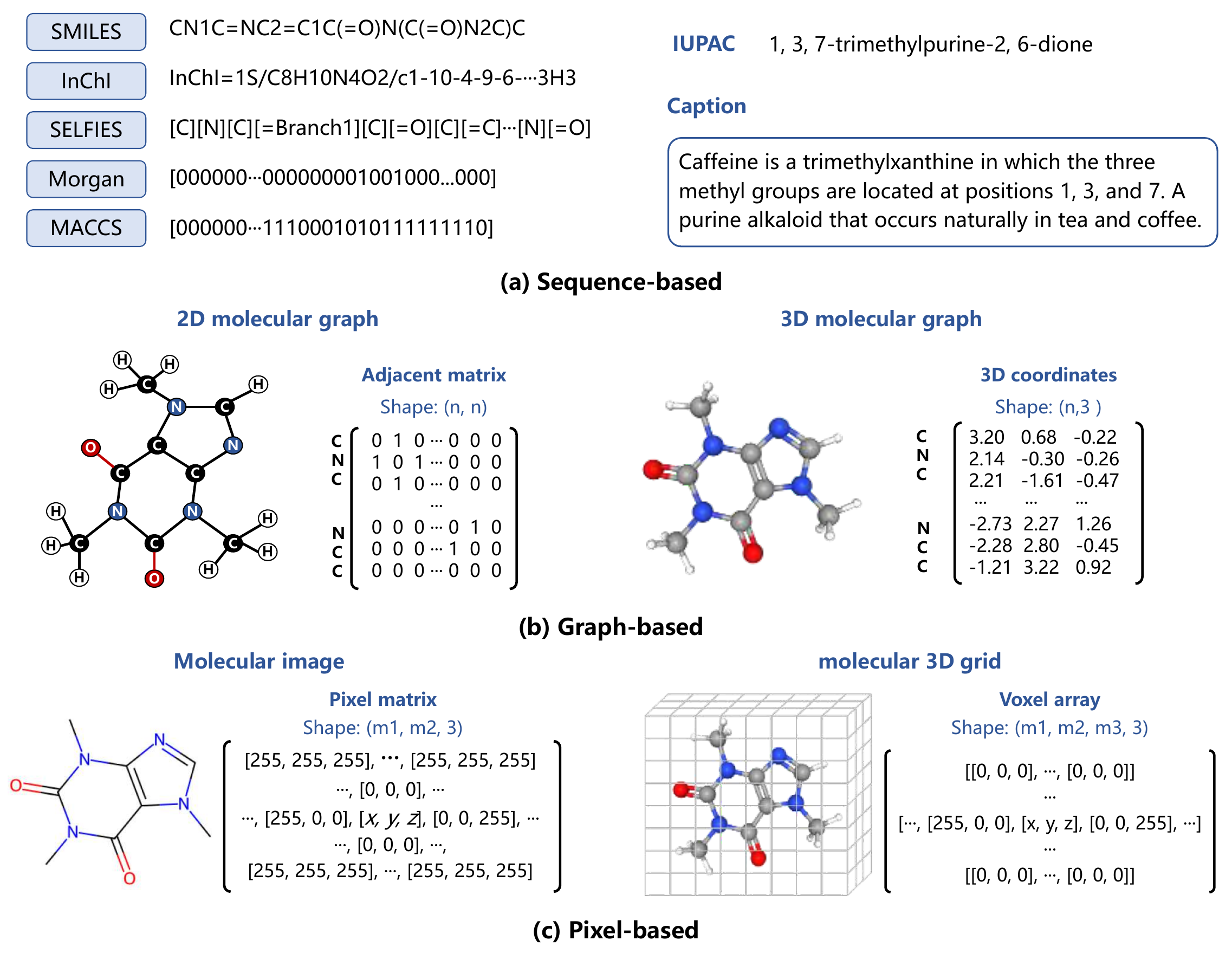}
    \caption{\textbf{Molecular modality:} 
    We illustrate the transformation of molecular modality crucial for MPP using the example of the caffeine molecule. This is demonstrated across three primary categories: sequence-based, graph-based, and pixel-based formats. Each format is derived from the SMILES representation of caffeine, using Python packages such as RDKit and software tools like PyMol. \textbf{a).} The sequence-based data section includes formats like SMILES and its variants (Canonical and Isomeric SMILES), molecular fingerprints (ECFP, Morgan, MACCS), and SELFIES, highlighting their roles in encoding molecular structures. \textbf{b).} Graph-based data represents caffeine as a graph with atoms as nodes and bonds as edges, enriched with 3D information for detailed structural insights. \textbf{c).} Pixel-based data showcases 2D images and 3D grids of caffeine, crucial for visual analysis and spatial interpretation. 
    }
    \label{fig:DF}
\end{figure*}

\noindent
In the field of molecular science, an vast range of molecular modality has been developed, each crucial for computational modeling and analysis. 
These formats are generally classified into three main types: text-based, graph-based, and pixel-based formats. 
Each type offers unique insights into molecular structures, contributing significantly to various aspects of molecular analysis. 
These diverse formats are illustrated in the Figure \ref{fig:DF}, which showcases the array of molecular modality available for DL methods.
\subsection{Sequence-based Data}
\noindent
Text-based formats are among the most commonly used representations in MPP due to their simplicity and efficiency. 
The most prominent of these is Simplified Molecular Input Line Entry System  (SMILES)\cite{weininger1988smiles}, which encodes molecules in linear strings, representing atoms and bonds in a compact, readable format. 
Variants of SMILES, such as Canonical SMILES\cite{weininger1989smiles} and Isomeric SMILES\cite{weininger1990smiles}, offer additional specificity, including stereochemistry information. 
Other notable text-based formats include molecular fingerprints like ECFP\cite{rogers2010extended}, Morgan, and MACCS\cite{durant2002reoptimization}, which encode the presence of certain molecular features, and Self-referencIng Embedded Strings (SELFIES)\cite{krenn2020self}, a newer format designed for robustness in machine learning applications.
Additionally, IUPAC\cite{mcnaught1997compendium} and InChI\cite{heller2015inchi} codes are vital text-based molecular representations.
IUPAC provides systematic chemical nomenclature for clear scientific communication, while InChI offers standardized textual identifiers for chemical substances. 
These formats facilitate various computational tasks, from database searching to the generation of novel molecules using AI.
\subsection{Graph-based Data}
\noindent
In drug discovery, graph-based representations, which depict atoms as nodes and bonds as edges, effectively capture molecular structures, making them ideal for analyzing both topological and relational aspects of molecules.
The method includes the use of a 2D adjacency matrix or a set of edges to outline atom connectivity. 
This representation can be enhanced with 3D information, such as bond lengths and atom positions, transforming it into a 3D graph. 
Incorporating a 3D atom distance matrix further enriches this model, offering a comprehensive view of the molecular spatial structure. 
Graph-based formats, including 2D and 3D molecular structures, are crucial in drug discovery for conducting detailed molecular analyses and enhancing the understanding of complex molecular behaviors.
\subsection{Pixel-based Data}
\noindent
Pixel-Based Molecular Data Formats, such as 2D images and 3D grids, are essential components of molecular property prediction. 
These formats, easily generated by tools like RDKit\cite{landrum2013rdkit} and PyMol\cite{delano2002pymol} for 2D images and Libmolgrid\cite{sunseri2020libmolgrid} for 3D grids, offer clear and comprehensible visual representations of molecular structures. 
This visual aspect allows for straightforward human interpretation, aiding in the recognition of molecular patterns and the understanding of spatial relationships in computational modeling. 
This clarity in visualization is crucial for effectively analyzing molecular geometries and interactions.
\section{Domain Knowledge}
\label{s:DomainKnowledge}
\begin{figure*}[t]
    \centering
    \includegraphics[width=1.0\linewidth]{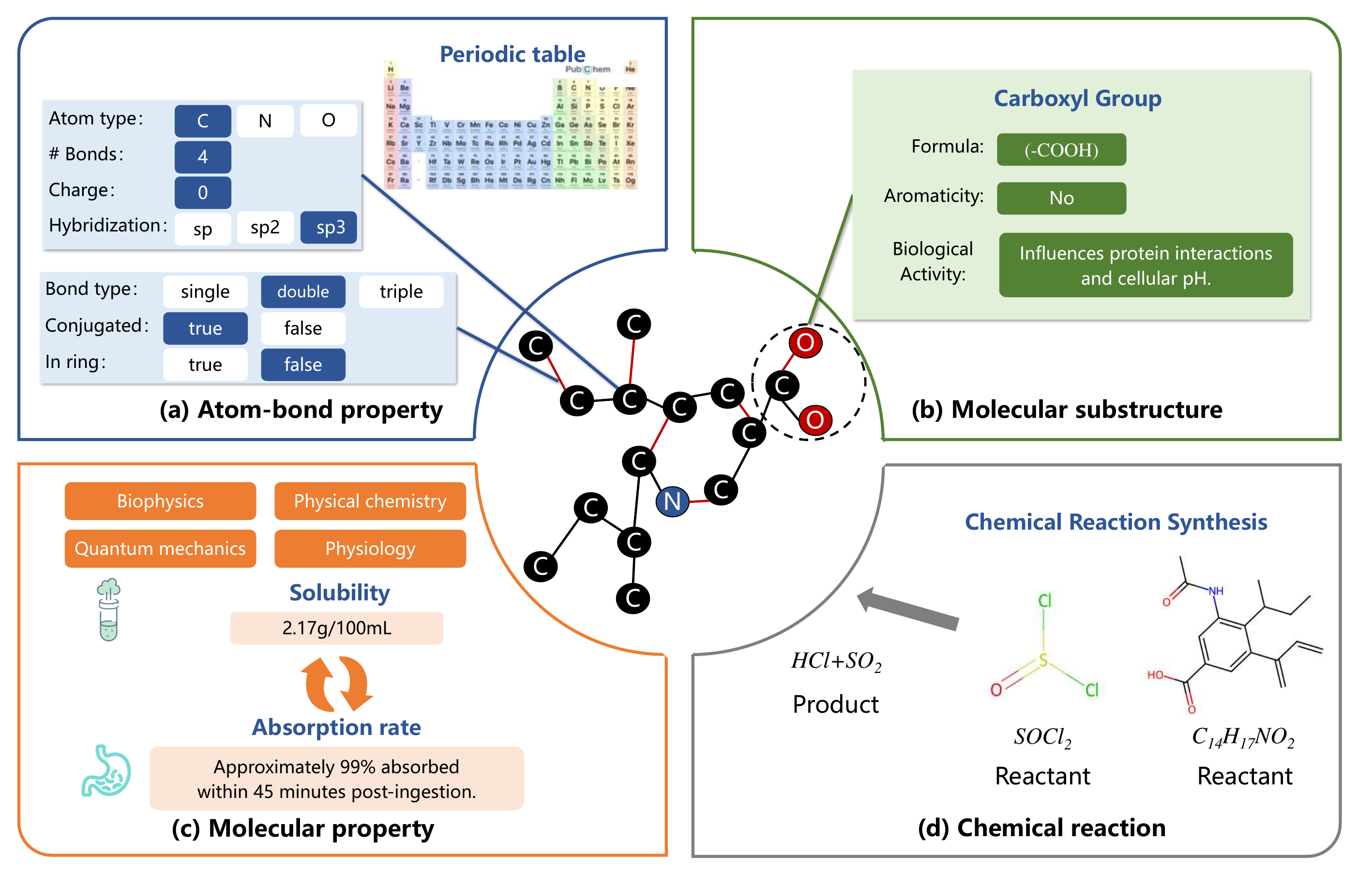}
    \caption{\textbf{Molecular domain knowledge:} 
    This figure categorizes molecular expert knowledge essential for MPP into four domains, using the molecule C=(CC(=C)c1cc(C(=O)O)cnc1C(C)CC as an example. \textbf{a).} In the atom-bond property section, we examine aspects such as the molecule’s atomic number, mass, valence, and bond types. \textbf{b).} Molecular substructure includes the functional groups, molecular fragments, and pharmacophores of this molecule, illustrating their influence on its chemical behavior and interactions. \textbf{c).} Molecular property covers a range of properties from quantum mechanics to physiology, showcasing how these properties affect the molecule's behavior in drug development. \textbf{d)}. Chemical reaction discusses the mechanisms of molecular transformations, highlighting the molecule's reactivity. 
    }
    \label{fig:DK}
\end{figure*}

\noindent
In molecular science, many domain knowledge from areas, like physics, chemistry, and biology, play a vital role.
This knowledge is methodically grouped into four key categories: atom-bond property, molecular substructure, chemical reactions, and molecular characteristics. 
Each category is integral for a comprehensive understanding and accurate interpretation of molecular data. 
The Figure \ref{fig:DK} showcases these categories in detail, providing an in-depth look at the essential aspects of molecular information interpretation.
\subsection{Atom-bond Property}
\noindent
In MPP, a deep understanding of atomic and bonding attributes is vital for accurately modeling molecular behaviors. 
Understanding atomic properties is essential for molecular analysis. For example, isotope numbers influence molecular weight and stability, and chirality is crucial for interactions and reactions within biological systems.
Hybridization types impact bonding patterns and molecular geometry.  
Atomic valence, number, mass, formal charge, and aromaticity significantly influence a molecule's chemistry.
Bond attributes like bond type and stereochemistry are key in determining molecular connectivity, reactivity, and shape, influencing interactions with biological targets. 
The direction and length of bonds also provide insights into spatial arrangement.
These detailed atomic and bond attributes collectively provide a comprehensive framework for molecular structure analysis, essential for effective predictive modeling in drug discovery.
\subsection{Molecular Substructure}
\noindent
In the realm of MPP, a deep comprehension of molecular substructures is indispensable. 
These substructures, including functional groups, molecular fragments, and pharmacophores, are fundamental in dictating the functions and interactions of molecule.\\
\indent These substructures, such as functional groups, molecular fragments, and pharmacophores, play key roles in understanding a molecule's behavior. 
Functional groups, such as hydroxyl (-OH) and carboxyl (-COOH), are specific group of atoms within a molecule that is responsible for the characteristic chemical reactions of that molecule, and are particularly influential in determining a molecule’s chemical behavior and interactions. 
For example, a hydroxyl group can significantly increase water solubility, thereby impacting a drug's absorption, distribution, and overall pharmacokinetics. 
\\
\indent Molecular fragments are larger portions of molecules, encompassing various structural elements like rings or chains.
Similarly, molecular fragments like benzene rings affect a molecule's stability and electronic, which in turn can alter its interaction with biological receptors or enzymes, impacting biological activity. 
Common molecular fragment methods are breaking of retrosynthetically interesting chemical substructure (BRICS)\cite{degen2008art}, Retrosynthetic Combinatorial Analysis Procedure (RECAP)\cite{lewell1998recap}, Murcko scaffolds\cite{bemis1996properties}, 
eMolFrags\cite{liu2017break}, and rdScaffoldNetwork\cite{kruger2020rdscaffoldnetwork}.\\
\indent A pharmacophore is an abstract representation of the molecular features that are necessary for a molecule to interact with a specific biological target to produce a desired biological effect. 
A pharmacophore with both a hydrogen bond donor and an acceptor in a specific spatial arrangement, for instance, can be crucial for binding to biological targets like enzymes or receptors, influencing the molecule’s effectiveness as a therapeutic agent. 
The accurate identification and understanding of these substructures are key to developing new pharmaceuticals, offering detailed insights into molecular interactions.
\subsection{Chemical Reaction}
\noindent
Chemical reactions involve the transformation of substances through the breaking and forming of chemical bonds, leading to the creation of new molecules with specific properties. 
For example, in the reaction C=CC(=C)c1cc(C(=O)O)cc(NC(=O)C)c1C(C)CC + SOCl2 → C=CC(=C)c1cc(C(=O)O)cnc1C(C)CC + HCl + SO2, the reactant interacts with thionyl chloride, resulting in a new product plus byproducts. 
This process highlights the role of reactants, products, and catalysts in affecting reaction outcomes and mechanisms. 
Such knowledge is vital for predicting reaction paths, designing new molecules with desired properties, and developing effective pharmaceuticals and novel compounds.
\subsection{Molecular Property}
\noindent
MPP in drug discovery is a multidisciplinary field, each discipline offering detailed insights into molecular behavior. 
Quantum mechanics, for example, delves into electronic properties like ionization potentials, crucial for understanding reaction mechanisms. 
Physical chemistry examines the stability of molecule, reactivity, and phase behaviors, impacting drug formulation. 
Biophysics explores molecular interactions within biological systems, crucial for drug-target binding studies. 
Physiology, on the other hand, assesses drug effects at an organismal level, influencing pharmacodynamics and pharmacokinetics. 
These interconnected properties, such as how a drug's solubility impacts absorption and bioavailability, highlight the need for a comprehensive understanding across levels, from atomic to organismal, to predict molecular properties accurately and develop effective pharmaceuticals. 
This integrative approach, encompassing everything from electron distribution to organismal response, is vital in the nuanced field of drug development.
\section{Modeling Method}
\noindent
Our paper provides a concise yet comprehensive examination of current DL methods in MPP. 
We first review molecular encoder architectures, and explore how these encoder align with the prior structural knowledge of molecules and how domain knowledge is integrated into them. 
Our review further emphasizes the utilization of unlabeled data, encompassing an exploration of self-supervised, semi-supervised, transfer learning, and multi-task learning strategies. 
To aid in understanding these intricate concepts, our paper includes detailed diagrams, which elucidate these advanced computational methods and their integration with fundamental molecular insights, thereby contributing to the advancement of MPP.
\subsection{Encoder}
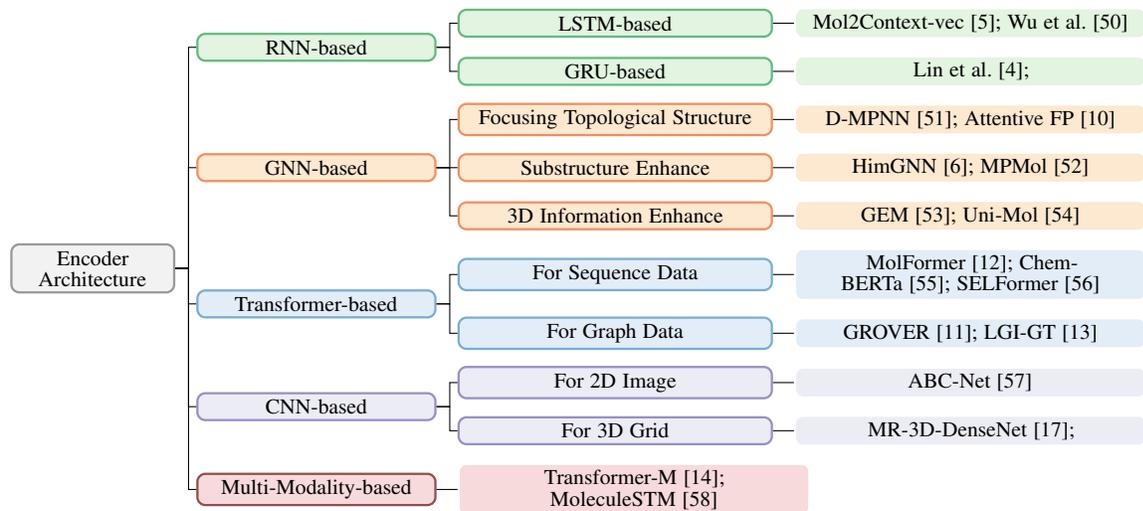
\begin{figure*}[!h]
    \scriptsize
    \hspace*{-30pt}
    \centering
    \begin{forest}
        for tree={
        forked edges,
        grow'=0,
        draw,
        rounded corners,
        node options={align=center,},
        text width=2.7cm,
        s sep=6pt,
        calign=edge midpoint,
        text=black,
        },
        [
            {\fontsize{8pt}{6pt}\selectfont Encoder~\\ Architecture },
            fill=gray!45, text=black, 
            parent
            [
                {\fontsize{8pt}{6pt}\selectfont RNN-based}, text=black,
                rnn
                [
                    {\fontsize{8pt}{6pt}\selectfont LSTM-based},
                    rnn_more
                    [
                        {\fontsize{8pt}{6pt}\selectfont Mol2Context-vec~\cite{lv2021mol2context}; Wu et al.~\cite{wu2021learning}},
                        rnn_work
                    ]
                ]
                [
                    {\fontsize{8pt}{6pt}\selectfont GRU-based},
                    rnn_more
                    [
                        {\fontsize{8pt}{6pt}\selectfont Lin et al.~\cite{lin2020novel};},
                        rnn_work
                    ]
                ]
            ]
            [
                {\fontsize{8pt}{6pt}\selectfont GNN-based}, text=black,
                gnn
                [
                    {\fontsize{8pt}{6pt}\selectfont Focusing Topological Structure},
                    gnn_more
                    [
                        {\fontsize{8pt}{6pt}\selectfont D-MPNN~\cite{yang2019analyzing}; Attentive FP~\cite{xiong2019pushing}},
                        gnn_work
                    ]
                ]
                [
                    {\fontsize{8pt}{6pt}\selectfont Substructure Enhance},
                    gnn_more
                    [
                        {\fontsize{8pt}{6pt}\selectfont HimGNN~\cite{han2023himgnn}; MPMol~\cite{ji2023metapath}},
                        gnn_work
                    ]
                ]
                [
                    {\fontsize{8pt}{6pt}\selectfont 3D Information Enhance},
                    gnn_more
                    [
                        {\fontsize{8pt}{6pt}\selectfont GEM~\cite{fang2022geometry}; Uni-Mol~\cite{zhou2023unimol}},
                        gnn_work
                    ]
                ]
            ]
            [
                {\fontsize{8pt}{6pt}\selectfont Transformer-based}, text=black,
                transformer
                [
                    {\fontsize{8pt}{6pt}\selectfont For Sequence Data},
                    transformer_more
                    [
                        {\fontsize{8pt}{6pt}\selectfont MolFormer~\cite{ross2022large}; ChemBERTa~\cite{chithrananda2020chemberta}; SELFormer~\cite{yuksel2023selformer}},
                        transformer_work
                    ]
                ]
                [
                    {\fontsize{8pt}{6pt}\selectfont For Graph Data},
                    transformer_more
                    [
                        {\fontsize{8pt}{6pt}\selectfont GROVER~\cite{rong2020self}; LGI-GT~\cite{yin2023lgi}},
                        transformer_work
                    ]
                ]
            ]
            [
                {\fontsize{8pt}{6pt}\selectfont CNN-based}, text=black,
                cnn
                [
                    {\fontsize{8pt}{6pt}\selectfont For 2D Image},
                    cnn_more
                    [
                        {\fontsize{8pt}{6pt}\selectfont ABC-Net~\cite{zhang2022abc}},
                        cnn_work
                    ]
                ]
                [
                    {\fontsize{8pt}{6pt}\selectfont For 3D Grid},
                    cnn_more
                    [
                        {\fontsize{8pt}{6pt}\selectfont MR-3D-DenseNet~\cite{liu2019multiresolution};},
                        cnn_work
                    ]
                ]
            ]
            [
                {\fontsize{8pt}{6pt}\selectfont Multi-Modality-based}, text=black,
                multi
                [
                    {\fontsize{8pt}{6pt}\selectfont Transformer-M~\cite{luo2022one}; MoleculeSTM~\cite{liu2023multi}},
                    multi_work
                ]
            ]
        ]
    \end{forest}
    \caption{
    \textbf{The molecular encoder method summary.} We categorize molecular encoder method into five types: RNN-based, GNN-based, Transformer-based, CNN-based, and Multi-Modality-based. For each category, key techniques and notable advancements utilized in various influential studies are highlighted, showcasing the evolution and diversification of approaches in molecular encoding.}
    \label{fig:EA_MM}
\end{figure*}

\begin{figure*}[!h]
    \centering
    \includegraphics[width=1.0\linewidth]{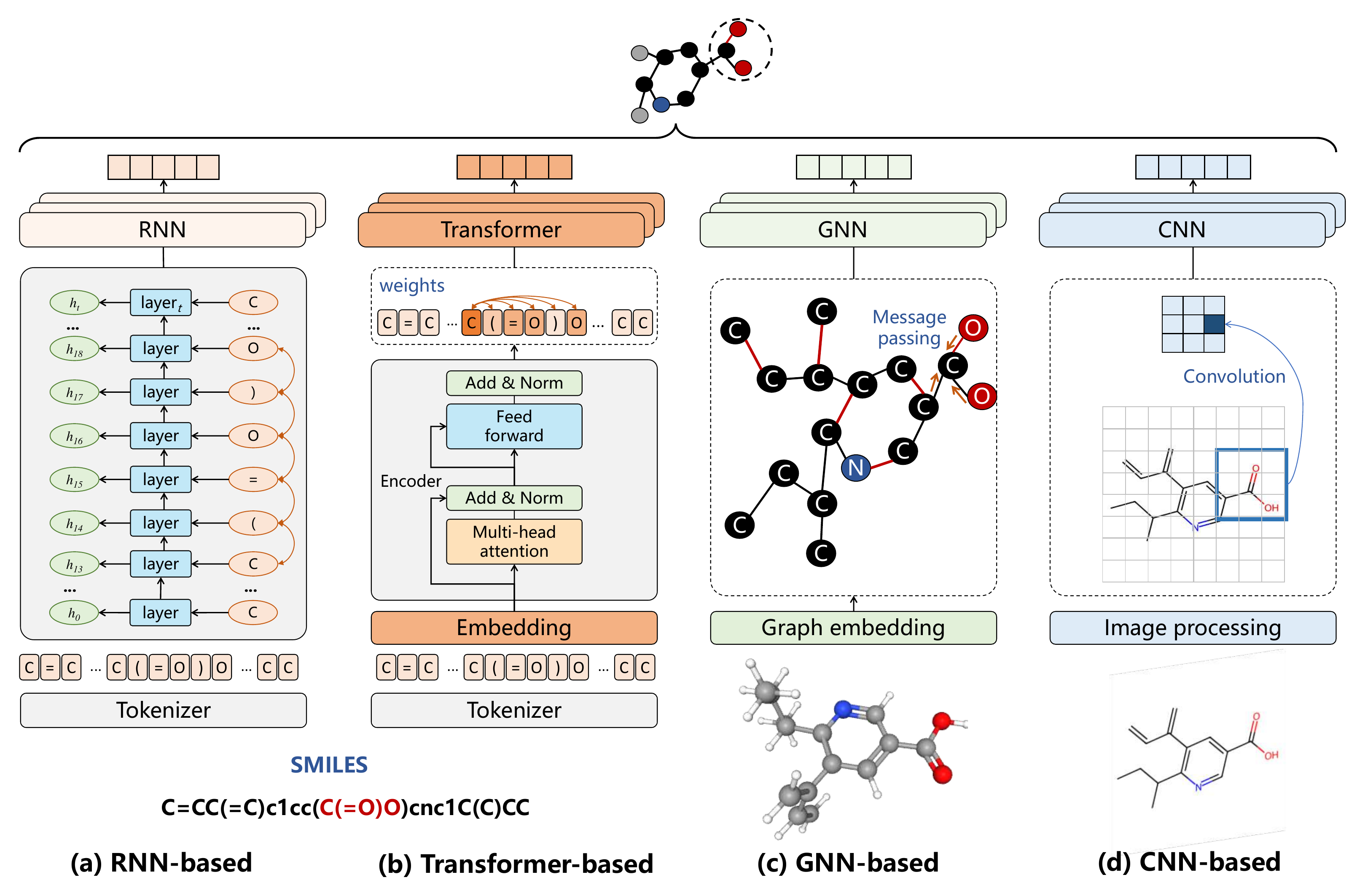}
    \caption{\textbf{Molecular encoder architectures:} 
    This figure categorizes molecular encoder architectures for single modality into four types: RNN-based, Transformer-based, GNN-based, and CNN-based. Each type is assessed for its ability to capture information about functional groups like carboxyl groups (-COOH) in the molecule C=(CC(=C)c1cc(C(=O)O)cnc1C(C)CC. \textbf{a).} RNN-based encoders process sequence data, maintaining a memory of previous inputs to effectively capture sequential patterns of (-COOH). \textbf{b).} Transformer-based models utilize Self-Attention mechanisms, enabling them to identify and focus on the (-COOH) group's specific interactions within the molecular sequence. \textbf{c).} GNN-based architectures employ a message passing strategy, extracting the topological information of (-COOH) within the molecule's graph structure. \textbf{d).} CNN-based models analyze spatial patterns through convolution layers, identifying sub-images that contain the (-COOH) group. This visualization highlights how each encoder type uniquely processes and interprets the molecular structure for MPP.
    }
    \label{fig:EA}
\end{figure*}

\noindent
In MPP, encoder architectures play a key role in transforming raw molecular data into meaningful representations. 
This section examines a variety of encoder architectures, each tailored to specific molecular modality and complexities. 
We categorize four main types of encoders for single data sources: RNN-based, GNN-based, Transformer-based, and CNN-based. 
Each type is analyzed for its alignment with molecular prior structural knowledge and the integration of domain-specific information. 
Additionally, we examine multi-modality based encoders, which handle multiple data sources, highlighting their unique characteristics, applications, and the challenges they address in molecular representation learning. 
The detailed aspects of these encoder architectures are illustrated in the Figure \ref{fig:EA_MM} and Figure \ref{fig:EA}.\\
\subsubsection{RNN-based}
\noindent
RNN, like long short-term memory (LSTM)\cite{liu2015multi} and gated recurrent unit (GRU)\cite{chung2015gated}, are adept at processing sequential data, with a unique internal memory feature that allows them to maintain context and order in sequences. This capability makes RNNs highly effective for tasks involving sequences data. 
Nowadays some work uses RNN-based model to analyze 1D molecular data, such as SMILES.\\ 
\indent Lin et al.\cite{lin2020novel} first transformed SMILES into sample vectors, which were then processed using bidirectional GRU neural networks to predict molecular properties, illustrating an innovative approach in training models for molecular property prediction.
Lv et al.\cite{lv2021mol2context} introduced Mol2Context-vec to address the challenge of representing molecular substructures and their polysemous nature, integrating different internal state levels for dynamic representations. 
To highlight the SMILES characters that are more important for the prediction tasks, Wu et al.\cite{wu2021learning} utilized the bidirectional long short term memory attention network in which they employed a novel multi-step attention mechanism to facilitate the extracting of key features from the SMILES strings. 
Nazarova et al.\cite{nazarova2021dielectric} used the single-layer Elman RNN to identify correlations between the structure of polymers of the norbornene class and their permittivity while using the SMILES notation in binary and decimal representations.
Wang et al.\cite{wang2019predictive} employs a Tree-structured LSTM network with signature descriptors to automatically generate expressive signatures for molecular structures, enabling the efficient representation of their structural information and connectivity in a single-step process.
\\
\indent These works demonstrate the effectiveness of RNN in extracting semantic information from SMILES sequences, paralleling methods in natural language processing(NLP). 
However, they face challenges when incorporating varied expert knowledge and managing long SMILES sequences, and the focus of RNN-based models on adjacent characters hampers effective interactions between distant atoms. 
This limitation can affect their ability to capture extensive structural relationships, especially when important atoms within the same functional group are distantly placed in the sequence.
\subsubsection{GNN-based}
\noindent
Molecules can be effectively represented as graphs, with atoms as nodes and chemical bonds as edges.
GNNs are well-suited to learn from this representation, utilizing layers that enable message passing.
In GNNs, node embeddings are updated by aggregating information from neighboring nodes, allowing the network to capture molecular features through atom-level interactions.
This method provides a detailed understanding of molecular structures by considering both individual atomic characteristics and their interconnections within the molecule.
Yang et al.\cite{yang2019analyzing} construct molecular encodings by using convolutions centered on bonds instead of atoms, thereby avoiding unnecessary loops during the message passing phase of the algorithm.
AttentiveFP\cite{xiong2019pushing} not only characterizes the atomic local environment by propagating node information from nearby nodes to more distant ones but also allows for nonlocal effects at the intramolecular level by applying a graph attention mechanism.
Withnall et al.\cite{withnall2020building} introduce attention and edge memory schemes to the existing message passing neural network framework.
To address insufficient bond information extraction, Li et al.\cite{li2021trimnet} explicitly drop the matrix mapping of edge features and employ a triplet message mechanism. This mechanism calculates messages from atom-bond-atom information and updates the hidden states of neural networks.
Zhang et al.\cite{zhang2022coatgin} propose CoAtGIN, which uses k-hop convolution to capture long-range neighbor information at the local level and utilizes linear attention to aggregate the global graph representation according to the importance of each node and edge at the global level.\\
\indent But these methods focus on atom (node) or bond (edge) information. To address this issue, Song et al.\cite{song2020communicative} propose a Communicative Message Passing Neural Network to improve molecular embedding by strengthening the message interactions between nodes and edges through a communicative kernel.
SC-NMP\cite{fan2021propagation} aggregates the node representations of the current step and the graph representation of the previous step, and proposes densely self-connected neural message passing, which connects each layer to every other layer in a feed-forward fashion.
To extract useful interactions between a target atom and its neighboring atomic groups, Li et al.\cite{li2021introducing} proposed a new graph learning paradigm based on a block design named block-based GNN and demonstrated that the network degradation problem can be reduced by applying a block design with normalization and skip-connection.
Ma et al.\cite{ma2020multi} employ cross-dependent message passing strategy to integrate the node-centered and edge-centered encoders.
Liu et al.\cite{liu2021hypergraph} develop a hypergraph-based topological framework to characterize detailed molecular structures and interactions at the atomic level. They have recently proposed embedding homology and persistent homology.
Feng et al.\cite{feng2022mgmae} transform each molecular graph into a heterogeneous atom-bond graph to fully utilize the bond attributes and design unidirectional position encoding for such graphs.
Biswas\cite{biswas2023predicting} pass additional atomic and molecular features, including 2D RDKit descriptors, Abraham parameters, QM descriptors, and 3D geometries, to improve the model performance.
Hasebe\cite{hasebe2021knowledge} proposed a knowledge-embedded message passing nerual network that can be supervised together with nonquantitative knowledge annotations by human experts on a chemical graph. 
This graph contains information on the important substructure of a molecule and its effect on the target property.
Yang et al.\cite{yang2021deep} extract physical information with a neural physical engine that learns molecular conformations by simulating molecular dynamics with parameterized forces. They then employ this physical information as supplementary data for predicting molecular properties.
\\
\indent However, most methods essentially attribute predictions to individual nodes, edges, or node features. This kind of interpretability is only partially compatible with chemists’ intuition at best. Chemists are more accustomed to comprehending the causal relationship between molecular structures and properties in terms of chemically meaningful substructures, such as functional groups, rather than individual atoms or bonds.
Zang et al.\cite{zang2023hierarchical} decompose the molecular graph by BRICS and additional decomposition to construct a motif-level graph, in which corresponding multi-level generative and predictive tasks are designed as self-supervised signals.
As the graph pooling technique for learning expressive graph-level representation is critical yet still challenging, Liu et al.\cite{liu2021hierarchical} propose master-orthogonal attention, a novel cross-level attention mechanism specifically designed for hierarchical graph pooling. 
To fully explore higher-order substructure information, Gao et al.\cite{gao2021higher} propose substructure interaction attention, which takes both the information of neighbors' substructures and the interaction information among them into account during the aggregation process.
To retain locality and linear network complexity, Bouritsas et al.\cite{bouritsas2022improving} employ a topologically-aware message passing scheme based on substructure encoding, which does not attempt to adhere to the Weisfeiler-Leman hierarchy.
Addressing the oversmoothing problem in multi-hop operations, Ye et al.\cite{ye2022molecular} construct a composite molecular representation with multi-substructural feature extraction and process such features effectively with a nested convolution plus readout scheme to capture interacting substructural information.
Zhu\cite{zhu2022hignn} utilize corepresentation learning of molecular graphs and chemically synthesizable BRICS fragments. Furthermore, a plug-and-play feature-wise attention block is first designed in the their model architecture to adaptively recalibrate atomic features after the message passing phase.
To accurately model the complex quantum interactions inherent in molecules, Lu et al.\cite{lu2019molecular} utilize a sophisticated hierarchical graph neural network, which directly extracts features from both the conformation and spatial information of molecules, and then integrates these features through multilevel interactions.
Fey et al.\cite{fey2020hierarchical} take in two complementary graph representations: the raw molecular graph representation and its associated junction tree, where nodes represent meaningful clusters in the original graph.
Focusing on the molecular hierarchical relationship, Han et al.\cite{han2023himgnn} propose a simple yet effective rescaling module, called contextual self-rescaling, that adaptively recalibrates molecular representations by explicitly modeling interdependencies between atom and motif features.
Ji et al.\cite{ji2023metapath} model a molecule as a heterogeneous graph and leverage metapaths to capture latent features for chemical functional groups. They also design a hierarchical attention strategy to aggregate heterogeneous information at both the node and relation levels.
To extract functional groups as motifs for small molecules, Wu et al.\cite{wu2023molformer} construct a heterogeneous molecular graph with both atom-level and motif-level nodes and adopt a heterogeneous self-attention layer to distinguish the interactions between multi-level nodes.
\\
\indent Since different 3D structures may lead to dissimilar molecular properties despite having the same 2D molecular topology, Recently, many works utilizing molecular 3D structures have been introduced.
To emphasize equivariant constraints, Fuchs et al.\cite{fuchs2020se} utilize the explicit increase of equivariance constraints in self-attention mechanisms.
As rotation-invariant representations struggle to convey directional information, Schutt et al.\cite{schutt2021equivariant} proposed rotationally equivariant message passing, exemplified by the Polarizable Atom Interaction Neural Network architecture. 
Brandstetter et al.\cite{brandstetter2021geometric} expand equivariant graph networks to include not only invariant scalar attributes but also covariant information like vectors or tensors. This model consists of steerable MLPs, capable of incorporating geometric and physical information within its message passing and update functions. 
Gasteiger et al.\cite{gasteiger2021gemnet} show the universality of spherical representations and employ a two-hop message passing mechanism with directed edge embeddings for rotationally equivariant predictions, and utilize symmetric message passing, augmented with geometric information, to enhance our model's efficacy in MPP.
Gasteiger et al.\cite{gasteiger2020fast} integrate directional information and interatomic distances by embedding and updating messages between atoms, using a spherical 2D Fourier-Bessel basis to jointly represent distances and angles.
To model angular relationships among neighboring atoms in a GNN, ensuring constraints like rotation invariance and energy conservation, Shuaibi et al.\cite{shuaibi2021rotation} utilize a per-edge local coordinate frame and innovate a spin convolution, thereby securing rotation invariance in edge messaging.
Fang et al.\cite{fang2022geometry} proposed a self-supervised framework using molecular geometric information by constructing a new bond angle graph, where the chemical bonds within a molecule are considered as nodes and the angle formed between two bonds is considered as the edge. 
\\
\indent The GNN-based model section concludes by recognizing that while GNN excel in capturing molecular topological information and integrating domain knowledge, their effectiveness is hindered by the small-world phenomenon. This characteristic leads to over-smoothing in deeper networks, where nodes lose feature distinctiveness, impacting predictive accuracy. Additionally, the specialized structure of GNN makes it challenging to scale up with increased parameters, limiting their capability to handle large molecular datasets effectively.
\subsubsection{Transformer-based}
\noindent
Originally excelling in NLP, the Transformer architecture is renowned for its self-attention mechanism, which allows for parallel processing of entire sequences. 
This capability enables it to efficiently manage long-range dependencies within data, making it highly effective in MPP. 
Its adeptness at understanding detailed contextual relationships enhances the accuracy and computational efficiency in predictive modeling. \\
\indent Wang et al.\cite{wang2019smiles} and Chithrananda et al.\cite{chithrananda2020chemberta} use Transformer to extract molecular information from SMILES, which is treated as natural language.
Wang et al.\cite{wang2021molcloze} proposed two significant advances in molecular data processing: structural fingerprint tokenization for more efficient molecule graph tokenization and normalized graph raw shortcut-connection to enhance latent representations in complex model structures. 
To address challenges in the validity and robustness of SMILES representations, Yüksel et al.\cite{yuksel2023selformer} uniquely utilizes SELFIES, a robust and flexible molecular representation format, to learn high-quality molecular features, enhancing the reliability of molecular data analysis in computational chemistry.
To predicting activity coefficients in binary mixtures, Winter et al.\cite{winter2022smile} integrate information from two SMILES strings representing the mixture components, along with temperature and token position data, into a unified matrix for input encoding. 
Ross et al.\cite{ross2022large} delved into the differences between absolute and relative position embeddings in SMILES representation, proposing an efficient linear attention approximation for the RoFormer\cite{su2024roformer} model, which focuses on relative positioning, to enhance molecular SMILES processing in deep learning applications.
\\
\indent The Transformer architecture, originally designed for sequence data, has been effectively adapted for molecular graph representation in recent research. Its proficiency in handling global molecular information enhances its utility in molecular property prediction, showcasing its versatility beyond traditional sequence analysis.
Maziarka et al.\cite{maziarka2020molecule} proposed the Molecule Attention Transformer, which adapts the Transformer architecture, augmenting the self-attention mechanism with inter-atomic distances and molecular graph structure.
Li et al.\cite{li2022kpgt} focus on chemical bonds in molecular representations, employing molecular line graphs to illustrate edge adjacencies in original molecular graphs. Each graph is augmented with a knowledge node containing molecular descriptors and fingerprints, connected to its original nodes
Rong et al.\cite{rong2020self} combines message passing networks with a Transformer-style architecture, extract vectors as queries, keys and values from nodes of the graph, then feed them into the attention block.
Park et al.\cite{park2022grpe} introduced Graph Relative Positional Encoding, which effectively encodes graph structures by concurrently addressing node-topology and node-edge interactions, bypassing the need for linearization.
Hussain et al.\cite{hussain2022global} developed the Edge-augmented Graph Transformer, employing global self-attention rather than traditional static convolutional aggregation. This design facilitates dynamic, long-range node interactions and incorporates edge channels for evolving structural information, enabling direct predictions on edges and links.
Masters et al.\cite{masters2022gps++} integrate a substantial message-passing module with a biased self-attention layer to facilitate both localized biases and broad-scale communication. 
Chen et al.\cite{chen2023graph} proposed Graph Propagation Attention, which explicitly handles node-to-node, node-to-edge, and edge-to-node interactions, allowing for comprehensive information propagation. 
Yin et al.\cite{yin2023lgi} developed a method that alternates between GNN and Transformer layers, repeated in sequence. This approach effectively blends local and global information, allowing the Graph Transformer to comprehensively integrate node data from both nearby and distant sources.
To extrace the coarse-grained view, Ren et al.\cite{ren2023enhancing} make the molecular graph first enters the message passing phase of the traditional GNN layers to update the node embeddings, then enters graph transformation layers to learn different granular information.
To achieve one encoder for extracting 2D or 3D information, Luo et al.\cite{luo2022one} use two separated channels to encode 2D and 3D structural information and incorporate them with the atom features in the network modules.
To fully leverages chemical knowledge, Gao et al.\cite{gao2023transfoxmol} construct an embedding unit comprising a GNN and a Transformer to balance the neighboring and distant interactions of an atom, and more attention is given to conjugated systems, unsaturated bonds, heteroatoms and the molecular topology.
To extract the molecular fragment information, Jiang et al.\cite{jiang2023pharmacophoric} design a pharmacophoric-constrained multi-views molecular representation graph, enabling PharmHGT to extract vital chemical information from functional substructures and chemical reactions.
\\
\indent Transformer have demonstrated effectiveness in recent work, particularly with sequence data like SMILES, where they treat it similarly to natural language. Their global information extraction capabilities also extend to molecular graph representation. Recent innovations combine Transformer with GNN, enabling simultaneous local and global data analysis. This blend showcases Transformer's strength in handling large molecular datasets and extracting comprehensive insights, vital in MPP.
\subsubsection{CNN-based}
\noindent
CNN, known for processing grid-like topology data, are adept at extracting features through convolutional layers and efficiently detecting local patterns. 
This makes them highly effective for image and pattern recognition tasks, a trait utilized extensively in MPP. \\
\indent To extract the local pattern of 1D molecular data, Hirohara et al.\cite{hirohara2018convolutional}'s innovative application of CNNs to SMILES data for chemical motif detection marked a significant step in computational drug discovery.
Chen et al.\cite{chen2021different} highlighted the impact of SMILES molecular enumeration on CNNs' performance in solubility prediction. \\
\indent As DL methods have achieved great success in the image processing field, some work used CNN to extracted 2D molecular image, but the size of the same atom/structure is vibrational in different molecules because of the fixed size of the whole molecular image.
To address this issue, Zhang et al.\cite{zhang2022abc} introduced ABC-Net, predicting graph structures by representing atoms and bonds as points, utilizing CNN-generated heat-maps. 
Jiang et al.\cite{jiang2022molecular} proposed an equal-sized molecular persistent spectral image, and encoder it with CNN model to extract molecular representation.\\
\indent As the visual representation of molecular structure, 3D molecular grid is important for extracting molecular 3D information.
However, a direct 3D representation of a molecule with atoms localized at voxels is too sparse, which leads to poor performance of the CNNs. 
To addrress this issue, Denis et al.\cite{kuzminykh20183d} present a novel approach where atoms are extended to fill other nearby voxels with a transformation based on the wave transform. 
Shuai et al.\cite{liu2019multiresolution} utilize an atom-centered gaussian density model for 3D molecular representation, which involves defining multiple channels for different spatial resolutions corresponding to each atom type. 
Sunseri et al.\cite{sunseri2020libmolgrid} facilitates the use of grid-based molecular representations in DL, generating 3D arrays of voxelized molecular data compatible with various DL frameworks.\\
\indent The research we have reviewed indicates that CNN-based networks excel at encoding pixel-based data, like 2D images and 3D grids, understandable to humans. 
This ability of CNN to efficiently extract local and global information from such data is essential for analyzing molecular behaviors.
\subsubsection{Multi-modality-based}
\noindent
Multi-modal learning, initially prominent in computer vision, is now widely applied in various fields for its ability to handle and integrate different data types. Its key benefit is enhancing model robustness by using complementary data sources. This approach has gained traction in molecular property prediction.
\\
\indent Due to the significant local chemical information contained in fingerprints may assist models to achieve superior results, Cai et al.\cite{cai2022fp} and Wang et al.\cite{wang2019molecule} termed fingerprints and graph neural networks, which combined and simultaneously learned information from molecular graphs and fingerprints for MPP.
Not only fingerprint, Liu et al.\cite{liu2023prediction}, MolFM\cite{luo2023molfm}, Sun et al.\cite{sun2022molecular}, GraSeq\cite{guo2020graseq} and GIT-Mol\cite{liu2024git} employ different encoders to process information from SMILES strings and molecular graphs, respectively.
Tang et al.\cite{tang2022merged} encode molecule by using molecular descriptors and fingerprints, molecular graph and SMILES text notation.
Liu et al.\cite{liu2023multi} combine molecular structural data and textual knowledge to enhance molecular comprehension, jointly learning the chemical structures of molecules and textual knowledge.
Zhang et al.\cite{zhang2023transg} use molecular mass spectrum as another representation to provide supplement information which is not contained in the graph data.
To address neglects 3D stereochemical information, Chen et al.\cite{chen2021algebraic} propose an algebraic graph-assisted bidirectional Transformer framework by fusing SMILES and algebraic graph representations. 
By broad learning of many molecular descriptors and fingerprint features, MolMap\cite{shen2021out} was developed for mapping these molecular descriptors and fingerprint features into robust two-dimensional feature maps. 
To integrate the 3D coordinates information, Zhou et al.\cite{zhou2023unimol} employ the atom distance matrix as the position encoding.
Liu et al.\cite{liu2022spherical} incorporates comprehensive relational data, including distance, angle, and torsion information between atoms, extending beyond the traditional edge-based 1-hop interactions.
Wang et al.\cite{wang2022advanced} embed both molecular graphs and sequences, then create a joint embedding space alongside modality-specific spaces to  ensure that the multi-modal data maintains both its distinctive characteristics and a consistent representation across different modalities.
\\
\indent 
In conclusion, the above work underscores the effectiveness of multi-modal learning in the context of MPP. This approach facilitates the seamless integration of various molecular modality, including sequences, graph data types, and molecular descriptors. By amalgamating these diverse sources of information, multi-modal learning provides a richer and more nuanced understanding of molecular properties, which is essential for achieving accurate predictions. 
\\
\subsection{Training Strategy}
\begin{figure*}[!h]
    \scriptsize
    \hspace*{-30pt}
    \centering
    \begin{forest}
        for tree={
        forked edges,
        grow'=0,
        draw,
        rounded corners,
        node options={align=center,},
        text width=2.7cm,
        s sep=6pt,
        calign=edge midpoint,
        text=black,
        },
        [
            {\fontsize{8pt}{6pt}\selectfont Training~\\ Strategy },
            fill=gray!45, text=black, 
            parent
            [
                {\fontsize{8pt}{6pt}\selectfont Self-Supervised Learning}, text=black,
                rnn
                [
                    {\fontsize{8pt}{6pt}\selectfont Contrastive Learning: Multi-View},
                    rnn_more_cl
                    [
                        {\fontsize{8pt}{6pt}\selectfont GraphMVP~\cite{liu2021pre}; DVMP~\cite{zhu2023dual}; Zhu et al.\cite{zhu2022unified}},
                        rnn_work_cl
                    ]
                ]
                [
                    {\fontsize{8pt}{6pt}\selectfont Contrastive Learning: Domain knowledge Boost},
                    rnn_more_cl
                    [
                        {\fontsize{8pt}{6pt}\selectfont KANO~\cite{fang2023knowledge}; MoCL~\cite{sun2021mocl}; iMolCLR~\cite{wang2022improving}},
                        rnn_work_cl
                    ]
                ]
                [
                    {\fontsize{8pt}{6pt}\selectfont Contrastive Learning: Masking Strategy},
                    rnn_more_cl
                    [
                        {\fontsize{8pt}{6pt}\selectfont MolCLR~\cite{wang2022molecular}; GraphCL~\cite{you2020graph}; ImageMol~\cite{zeng2022accurate}},
                        rnn_work_cl
                    ]
                ]
                [
                    {\fontsize{8pt}{6pt}\selectfont Encoder-Recovery/Prediction},
                    rnn_more_ot
                    [
                        {\fontsize{8pt}{6pt}\selectfont KPGT~\cite{li2023knowledge}; Mole-BERT~\cite{xia2022mole}; K-BERT~\cite{wu2022knowledge}},
                        rnn_work_ot
                    ]
                ]
                [
                    {\fontsize{8pt}{6pt}\selectfont Substructure Enhance},
                    rnn_more_ot
                    [
                        {\fontsize{8pt}{6pt}\selectfont SME~\cite{wu2023chemistry}; FragCL~\cite{kim2023fragment}; HiMol~\cite{zang2023hierarchical}},
                        rnn_work_ot
                    ]
                ]
            ]
            [
                {\fontsize{8pt}{6pt}\selectfont Semi-Supervised Learning}, text=black,
                gnn
                [
                    {\fontsize{8pt}{6pt}\selectfont Consistency Regularization},
                    gnn_more
                    [
                        {\fontsize{8pt}{6pt}\selectfont InfoGraph*~\cite{sun2019infograph}; DropConn~\cite{zhang2023dropconn}},
                        gnn_work
                    ]
                ]
                [
                    {\fontsize{8pt}{6pt}\selectfont Pseudo Label},
                    gnn_more
                    [
                        {\fontsize{8pt}{6pt}\selectfont ASGN~\cite{hao2020asgn}; InstructBio~\cite{wu2023instructbio}},
                        gnn_work
                    ]
                ]
            ]
            [
                {\fontsize{8pt}{6pt}\selectfont Transfer Learning}, text=black,
                transformer
                [
                    {\fontsize{8pt}{6pt}\selectfont Property-Molecule Relation Enhance},
                    transformer_more
                    [
                        {\fontsize{8pt}{6pt}\selectfont Meta-GAT~\cite{lv2023meta}; GS-Meta~\cite{zhuang2023graph}},
                        transformer_work
                    ]
                ]
            ]
            [
                {\fontsize{8pt}{6pt}\selectfont Multi-Task Learning}, text=black,
                cnn
                [
                    {\fontsize{8pt}{6pt}\selectfont Biswas et al.~\cite{biswas2023predicting}},
                    cnn_work
                ]
            ]
        ]
    \end{forest}
    \caption{
    \textbf{The training strategy summary.} We categorize training strategies into four key types: Self-Supervised Learning, Semi-Supervised Learning, Transfer Learning, and Multi-Task Learning. Each category includes a detailed description of the main focuses and considerations prevalent in renowned studies, illustrating the diverse approaches and priorities within each training strategy for optimizing molecular property prediction.}
    \label{fig:TS_MM}
\end{figure*}
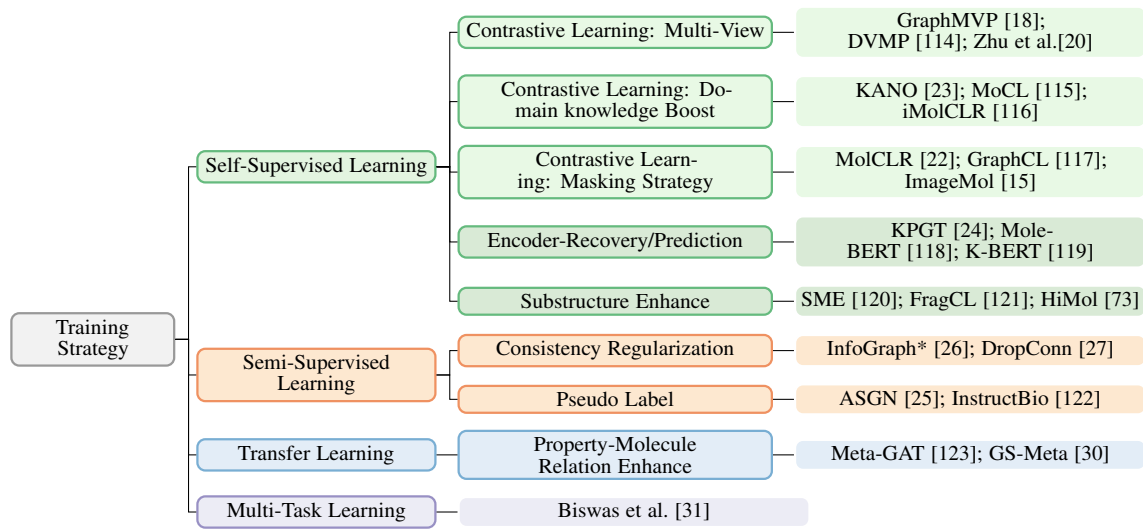

\begin{figure*}[t]
    \centering
    \includegraphics[width=1.0\linewidth]{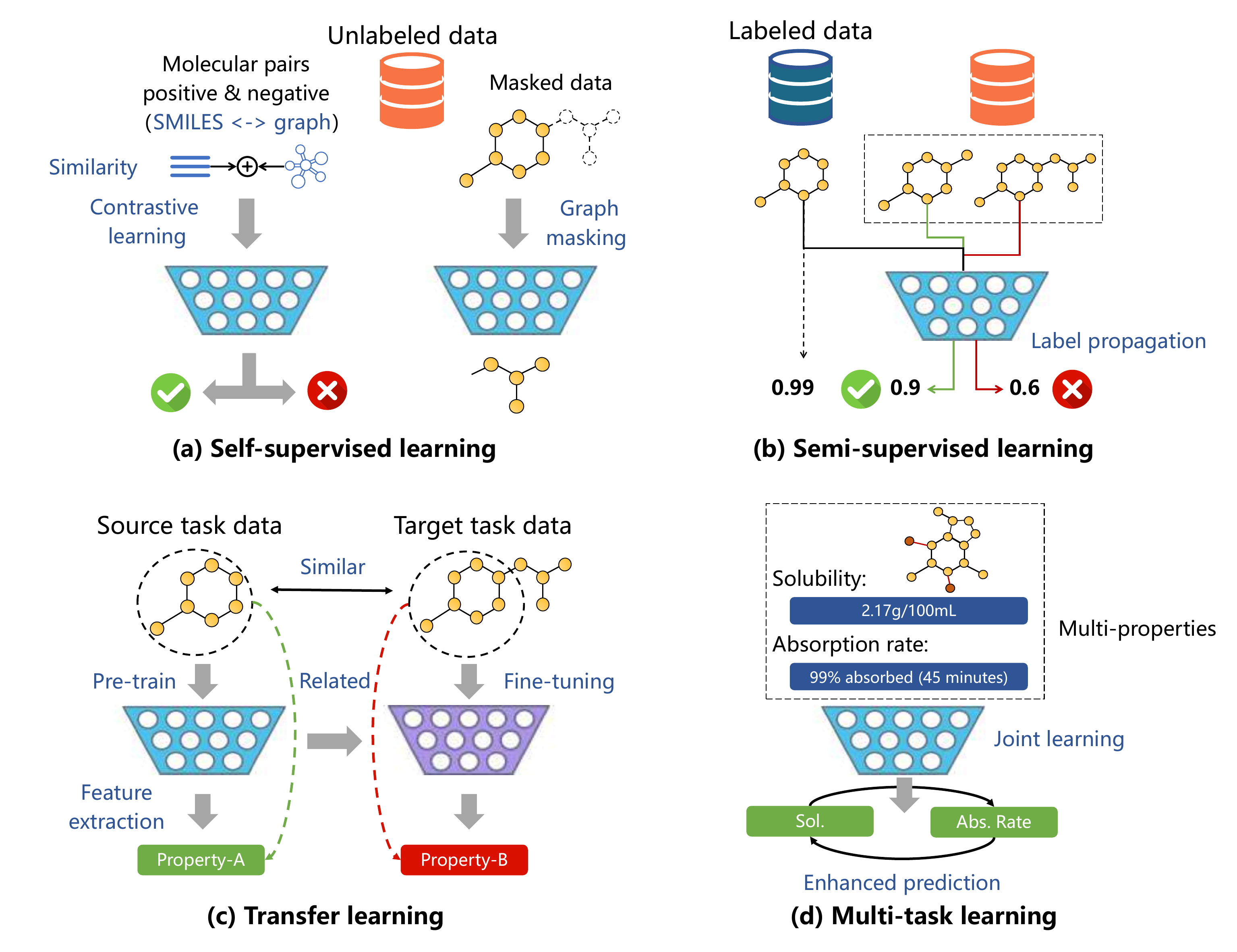}
    \caption{\textbf{Training strategy:} 
    The high cost of experiments often results in a scarcity of labeled data, leading to challenges like overfitting and poor generalization. To address this, our figure illustrates various advanced training strategies. \textbf{a).} Self-supervised learning, such as contrastive learning and masking recovery, utilizes unlabeled data for pretraining encoders, enabling them to learn molecular domain knowledge effectively. \textbf{b).} Semi-supervised learning methods like pseudo-labeling and co-training leverage both labeled and unlabeled data, enhancing the encoder's understanding of data distribution. \textbf{c).} Transfer learning strategy capitalizes on pretrained models from source tasks to boost performance on target tasks. \textbf{d).} Multi-task learning approaches combine related datasets to predict multiple properties simultaneously, benefiting from the relational aspects of molecular data.
    }
    \label{fig:TS}
\end{figure*}

\noindent
In this section, we introduce all approaches used to train DL models. While supervised learning has been traditionally predominant, its reliance on scarce labeled data presents limitations. 
To circumvent this, recent approaches have shifted towards unsupervised, self-supervised, and semi-supervised learning methods, capitalizing on the abundance of unlabeled data. 
Transfer learning is also employed to utilize models pretrained on unrelated data, enhancing the model's performance on specific tasks. 
Additionally, multi-task learning strategies are adopted to leverage related labeled data, further refining the model's accuracy in predicting molecular properties.
As Figure \ref{fig:TS_MM} and Figure \ref{fig:TS} shown, the details of training strategy are as followings.
\subsubsection{Self-supervised Learning}
\noindent
Self-supervised learning, widely used in NLP\cite{devlinBERTPretrainingDeep2018, floridi2020gpt}, utilizes unlabeled data to extract prior knowledge, proving effective in addressing labeled data scarcity. 
This method empowers models to learn comprehensive representations from abundant unlabeled data, enhancing their learning capabilities and insight extraction.\\
\indent 
Inspired by NLP, Wang et al.\cite{wang2019smiles}, Chithrananda et al.\cite{chithrananda2020chemberta}, Zhang et al.\cite{zhang2021mg}, Ahmad et al.\cite{ahmad2022chemberta} and Irwin et al.\cite{irwin2022chemformer} employ masked language modeling(MLM) on large scale unlabeled data to generate context-sensitive representation, treating SMILES as natural language. 
Ma et al.\cite{ma2020improving} use auto-encoder strategy in pretrain stage, first convert SMILES to a vector representation and then reconstructed representation back to SMILES to update the network. Furthermore,  Guo et al.\cite{guo2020graseq} fusion the molecular graph and SMILES representation to recontruct the SMILES. 
Except using SMILES as input, Yuksel et al.\cite{yuksel2023selformer} employs MLM in SELFIES representations in order to obtain their concise, flexible, and meaningful representations.
To let nodes appearing in similar structural contexts to nearby embeddings, Hu et al.\cite{hu2019strategies} propose a context prediction task by using subgraphs to predict their surrounding graph structures.
To address GNN oversmoothing and encourage latent node diversity, Godwin et al.\cite{godwin2021simple} employ denoise technique in which they corrupt the input graph with noise, and add a noise correcting node-level loss.
Zeng et al.\cite{zeng2022accurate} implemented an auto-encoder for molecular image reconstruction, using a discriminator to distinguish between real and fake molecular images. 
To expand atom vocabulary, Xia et al.\cite{xia2022mole} use a context-aware tokenizer to encode atom attributes into meaningful discrete codes, then randomly masking and recovering these codes to efficiently pretrain their encoder.
Intrinsically, for molecules, a more natural representation is based on their 3D geometric structures, which largely determine the corresponding physical and chemical properties.
To overcome the challenge of attaining the coordinate denoising objective, Liu et al.\cite{liu2022molecular} employ an SE(3)-invariant score matching strategy to successfully transform such objective into the denoising of pairwise atomic distances.
To capture the anisotropic characteristic of molecules, Feng et al.\cite{feng2023fractional} propose a novel hybrid noise strategy, including noises on both dihedral angel and coordinate, and also decouple the two types of noise and design a novel fractional denoising method, which only denoises the latter coordinate part.
For effectively learning 3D spatial representation, Zhou et al.\cite{zhou2023unimol} employ 3D position recovery and masked atom prediction as pretrain task. Further more, Jiao et al.\cite{jiao2023energy} exploit the Riemann-Gaussian distribution to ensure the loss to be E(3)-invariant, enabling more robustness.
To guild by the molecular domain knowledge and extract chemical information like chemists, Li et al.\cite{li2022kpgt, li2023knowledge} leverages the molecular descriptors and fingerprints, which serves as the semantics lost in the masked graph to guide the prediction of the masked nodes, thus making the model capture the abundant structural and semantic information from large-scale unlabeled molecules.
Wu et al.\cite{wu2022knowledge} proposed atom property prediction to discern finer differences between atoms, and MACCS fingerprints prediction, enabling their model to extract and learn predefined molecular features.
Gao et al.\cite{gao2022supervised} use atom charges and 3D geometries as inputs, with molecular energies as the target labels, aiming to effectively leverage energy information for enhanced molecular analysis.
To optimize multi-task integration and avoid ineffective transfer, Wang et al.\cite{wang2023automated} introduce a fusion strategy that utilizes a surrogate metric based on the total energy of all atoms in a molecule during the pretraining stage. 
Zang et al.\cite{zang2023hierarchical} designs three generative tasks that predict bond links, atom types, and bond types with the atom representations and designs two predictive tasks that predict the number of atoms and bonds with the molecule representation.
Zeng et al.\cite{zeng2023molkd} and Broberg et al.\cite{broberg2022pre} have developed methods to predict the product molecular SMILES based on the reactant molecular embedding. This approach allows for the extraction of chemical information from chemical reactions, providing insights into the molecular transformations involved in the reaction process.
\\
\indent Contrastive learning, a method distinguishing positive and negative molecule pairs, has become a key strategy in encoder pretraining for its ability to enhance molecular structure discernment. This technique is extensively utilized in numerous studies, making it a cornerstone for improving molecular structure recognition in various models. 
For the SMILES augmentation, Wu et al.\cite{wu2021learning} and Zhang et al.\cite{zhang2022pushing} implemented SMILES enumeration, a technique that varies starting atoms and traversal orders to represent a molecule with different SMILES, thereby uncovering more intricate patterns from complex SMILES structures.
Wu et al.\cite{wu2022knowledge} and Abdel et al.\cite{abdel2022large} utilized SMILES permutation as a data augmentation technique, involving the rearrangement of atoms in a SMILES string to create different representations without altering the underlying molecular structure.
For molecular graph augmentation, techniques like node dropping, edge perturbation, attribute masking, and subgraph masking are commonly used\cite{you2020graph, wang2022molecular, zheng2023casangcl, guan2023t}. However, these random masking methods may not effectively guide the encoder to identify the most crucial chemical information, and might result in the creation of less accurate positive and negative molecule pairs for the training process.
To capture important molecular structure and higher order semantic information, Liu et al.\cite{liu2022attention} adopted the graph attention network as the molecular graph encoder, and leveraged the learned attention weights as masking guidance to generate molecular augmentation graphs.
Lin et al.\cite{lin2022prototypical} first models the underlying semantic structure of the graph data via clustering semantically similar graphs to select the positive and negative pair and then reweights its negative samples based on the distance between their prototypes and the query prototype such that those negatives having moderate prototype distance enjoy relatively large weights.
Cui et al.\cite{cui2023mocgcl} utilize the GNN encoder and its momentum-update version\cite{he2020momentum} to generate positive samples at the representation level, and select the negative pairs by the semantic importance of nodes, which is calculated by eigenvector centrality iteration\cite{zaki2014data}.
Wang et al.\cite{wang2021molecular} employ a generative probabilistic model to learn molecular graph structures for topology augmentations and simultaneously develop feature selectors to mask less critical atom features, thus generating effective attribute-level augmentations.
To gain deeper insights into chemical information, many researchers incorporate domain knowledge into their contrastive learning approaches.
By using backbone and side-chain information, Liu et al.\cite{liu2022hiermrl} employ side-chain repetition, side-chain generation, backbone disruption, and backbone disruption + side-chain deletion strategy to generate hard positive, soft positive, soft negative and hard negative samples, respectively.
Sun et al.\cite{sun2021mocl} replaced a valid substructure by a bioisostere that introduces variation without altering the molecular properties too much, and treats them as positive pairs. Also, they optimize the similarity of molecule pairs embedding to be close to the similarity of their ECFP.
To avoid faulty negative pairs, Wang et al.\cite{wang2022improving} mitigate negative contrastive instances by considering ECFP similarities between molecule pairs.
Wang et al.\cite{wang2023molecular} calculate the weight vector using the self-attention mechanism to determine the selection probability of each character in SMILES and generate positive samples using three masked strategies: roulette masking, top masking, and random masking.
To maintain semantics between conformers, Moon et al.\cite{moon20233d} randomly selects molecules from the conformer pool instead of selecting the most stable molecules to learn the 3D structure abundantly.
Kuang te al.\cite{kuang20233d} consider conformations with the same SMILES as positive pairs and the opposites as negative pairs, while keeping the weight to indicate the 3D conformation descriptor and fingerprint similarity.
Knowledge graph (KG) is a semantic network composed of entities and their relations in the real world.\cite{wu2023medical}
Hua et al.\cite{hua2022chemical} use the atoms in SMILES as indices to query the embedding matrix to obtain entity and relation embeddings. For the entity and relation vectors of different atoms, they obtain the entity and relation embeddings of the SMILES through linear mapping, and finally concatenate the two vectors to obtain the final embedding representation.
Fang et al.\cite{fang2022molecular} first construct a Chemical Element KG based on periodic table of elements, to describes the relations between elements and their basic chemical attributes, Furthermore, they\cite{fang2023knowledge} construct another chemical Element KG based on the periodic table and Wikipedia pages to summarize the basic knowledge of elements and the closely related substructure. Those KG offers a comprehensive and standardized view from a chemical element perspective, and help to augment the original molecular graph with the guidance of KG.\\
\indent In the realm of MPP, a deep understanding of molecular substructures is increasingly recognized as crucial. Many recent studies leverage this domain knowledge to effectively identify and analyze important substructural information, significantly enhancing the understanding of molecular behavior.
Xu et al.\cite{xu2021self} aimed to preserve local similarities between graph instances by aligning embeddings of related subgraphs and differentiating these from unrelated pairs. They also implemented hierarchical prototypes to represent the latent distribution of graph datasets, enhancing data likelihood with respect to both GNN parameters and these hierarchical structures.
Wang et al.\cite{wang2022improving} employ BRICS to decompose different substructures which are considered as contrastive negative pairs.
Motifs, including chemical functional groups or fragments, serve as self-generated labels determined by their presence or absence in the graph. Shen et al.\cite{shen2020molgnn} and Rong et al.\cite{rong2020self} used these labels to pretrain their encoder.
To learn the local semantics, Luo et al.\cite{luo2022clear} use graph clustering techniques to partition each whole graph into several subgraphs while preserving as much semantic information as possible, and treat the molecular graph and the clustering graph as postive pair.
Benjamin et al.\cite{benjamin2022graph} extract substructure information by setting the junction tree(through a tree decomposition algorithm) reconstruction and fingerprint prediction task.
To analyze molecular GNN strictly in terms of chemically meaningful fragments, Wu et al.\cite{wu2023chemistry} identifies the most crucial set of substructures(BRICS and Murckoand functional groups) in a molecule that are responsible for a model’s prediction. 
HeGCL\cite{shi2023hegcl} introduce the meta-path view that provides semantic information, and encodes graph embeddings by maximizing mutual information between global and semantic representations obtained from the outline and meta-path view, respectively.
Hierarchical Molecular Graph is a usually way to extract the substructure molecular representations.
Zhu et al.\cite{zhu2022hignn} extracts hierarchical information by utilizing co-representation learning of molecular graphs and chemically synthesizable BRICS fragments, and also uses a feature-wise attention block to adaptively recalibrate atomic features after the message passing phase.
Kim et al.\cite{kim2023fragment} construct a bag of fragments from a molecule through fragmentation, treating a complete or incomplete bag as a positive or negative view of the original molecule, respectively.
Xie et al.\cite{xie2023self} proposed a fragment-based molecular graph (FMG) to represent the topological relationship between chemistry-aware substructures within a molecule. 
They then pretrained it on a fragment level using contrastive learning with well-designed hard negative pairs to extract node representations in FMGs.
Ji et al.\cite{ji2022relmole} decompose the molecular graph using a more reasonable method to construct the fragment graph. They select positive/negative pairs based on similarities between two-level molecule pairs and employ a contrastive loss function, as proposed by Hadsell et al.\cite{hadsell2006dimensionality}, to pretrain the encoder.\\
\indent Diverse data formats have been shown to be crucial for MPP, and the multi-modal approach, merging these formats, enhances prediction accuracy by offering a holistic view of molecules. This technique, increasingly adopted in research, combines different data types for a more detailed molecular analysis.
To leverage two popular molecular representations and augmentations for each modality, Pinheiro et al.\cite{pinheiro2022smiclr}, Zhang et al.\cite{zhang2022pseudo}, Zhu et al.\cite{zhu2023dual} and Sun et al.\cite{sun2022molecular} exploit two molecular representations that can be easily acquired from chemical space: the SMILES string and the molecular graph, and then make them as positive pairs.
Li et al.\cite{li2022geomgcl} utilized self-supervised learning by exploiting the relationship and consistency between 2D topological and 3D geometric structures of molecules. Additionally, Liu et al.\cite{liu2021pre} applied a generative self-supervised learning approach that focuses on intra-data knowledge, reconstructing key features at the individual data point level to enhance the understanding of molecular structures.
3D Infomax\cite{stark20223d} maximized the mutual information between learned 3D summary vectors and the representations of a GNN.
Zhu et al.\cite{zhu2022unified} implemented a multifaceted pretraining strategy involving the reconstruction of masked atoms and coordinates, generating 3D conformations based on 2D graphs, and creating 2D graphs from 3D conformations. 
Kim et al.\cite{kim2023fragment} focused on extracting explicit 3D geometric information by proposing a solution for predicting torsional angles between adjacent molecular fragments, thereby enhancing the depth and accuracy of 3D molecular analysis.
Zhu et al.\cite{zhu2024molecular} aimed to integrate multiple molecular feature views, including 2D and 3D graphs, Morgan fingerprints, and SMILES strings, ensuring cohesive embedding consistency between these among representations for a more unified molecular analysis.\\
\indent The reviewed works show that self-supervised learning, particularly through methods like encoder-recovery and contrastive learning, effectively utilizes unlabeled data to improve model generalization in MPP. 
These methods excel in learning prior knowledge through various pretraining tasks, allowing for integration of multi-modal data and domain knowledge. 
This approach significantly enhances the adaptability and performance of models in molecular property prediction scenarios.
\\
\subsubsection{Semi-supervised Learning}
\noindent
Semi-supervised learning effectively alleviates the scarcity of labeled molecular data in fields like MPP. By blending a small subset of labeled data with a larger pool of unlabeled data, it bridges the gap between fully supervised and unsupervised learning methods. \\
\indent Consistency regularization is based on the idea that applying realistic perturbations to unlabeled data should not significantly alter predictions, ensuring stability and reliability in the learning process.
InfoGraph*\cite{sun2019infograph} employ Mean-Teacher method\cite{tarvainen2017mean} to maximizes the mutual information between unsupervised graph representations and the representations learned by existing supervised methods in semi-supervised scenarios.
Chen et al.\cite{chen2021chemical} predict chemical toxicity and trained the network by the Mean Teacher SSL algorithm, which update the weights in teacher model by applying the Exponential Moving Average.
Zhang et al.\cite{zhang2023dropconn} propose a data augmentation which constructing new adjacency matrix and randomly masking the edges, and calculate the average of all data augmentation distributions and then employ MixMatch\cite{berthelot2019mixmatch} label guessing and sharpening method to minimize entropy and accurately guess labels based on the label distribution center.\\
\indent Proxy-label strategy, assigning temporary labels to unlabeled data, expand the training dataset when labeled data is limited. This approach enhances the model's learning process, with the proxy labels being iteratively refined for improved accuracy and generalization.
ASGN\cite{hao2020asgn} adopts a teacher-student framework to jointly exploit information from molecular structure and molecular distribution to learn general representation, then employs the active learning strategy in terms of molecular diversities to select informative data.
Yu et al.\cite{yu2020semi} have developed a semi-supervised drug embedding model that combines unsupervised learning from the chemical structures of drugs and drug-like molecules with supervised learning based on hierarchical relations from an expert-crafted drug hierarchy. This approach ensures a robust and comprehensive representation of drug properties.
Ma et al.\cite{ma2022robust} employ teacher-student framework, which use several epochs as a iteration, updating teacher model by the best student model. As the cross-entropy (CE) loss function is not proved to be robust to label noise during the training, they employ generalized CE\cite{zhang2018generalized} loss to boost the self-training.
To address data imbalance, Liu et al.\cite{liu2023semi} analyze the distribution of imbalanced annotated data and identify label ranges needing adjustment, and then use high-quality pseudo-labels create graph examples to augment under-represented areas, striving for an ideal balance in training data.
Wu et al.\cite{wu2023instructbio} introduces an instructor model to provide the confidence ratios as the measurement of pseudo-labels’ reliability. These confidence scores then guide the target model to pay distinct attention to different data points, avoiding the over-reliance on labeled data and the negative influence of incorrect pseudo-annotations.
\\
\indent This approach not only enhances model performance by utilizing the comprehensive information available in unlabeled data but also addresses the challenge of acquiring extensive labeled datasets, which is a common issue in MPP.\\
\subsubsection{Transfer Learning}
\noindent
Transfer learning strategies, widely adopted in various fields to address data scarcity, focus on enhancing prediction performance for tasks with limited data.\cite{zamir2018taskonomy, li2019deepchemstable, chen2021exploring} 
These strategies involve transferring knowledge from a data-rich source task to improve molecular representation learning ability in a data-scarce target task. Recently, there has been a significant increase in methods employing transfer learning, showcasing its growing importance and application across different domains.\\
\indent Sun et al.\cite{sun2022pemp} enhanced chemical and physiological property predictions by applying transfer learning, integrating insights from physics and physical chemistry to improve training outcomes.
Li et al.\cite{li2022improving} developed a framework for accurately estimating task similarity, which, as demonstrated in comprehensive tests, provides valuable guidance for enhancing the prediction performance of transfer learning in molecular property analysis.\\
\indent Meta-learning, focusing on rapid adaptation to new tasks with minimal data, is effective in addressing the lack of labeled molecular data. 
Many recent works\cite{ju2023few, nguyen2020meta, torres2023few, de2022graph, ham2023evidential} based on Model-Agnostic Meta-Learning (MAML), enabling rapid adaptation and learning in data-limited scenarios.
To effectively utilize correlations of molecules and properties, Lv et al.\cite{lv2023meta} construct a molecule-property relation Graph, where nodes represent molecules and properties connected by property labels, and then redefine a meta-learning episode as a subgraph within it, containing a target property node along with related molecule and auxiliary property nodes.
Chen et al.\cite{chen2022meta} developed ADKF-IFT, a model that separately trains a subset of parameters with meta-learning loss and adapts others using maximum marginal likelihood for each task. This method, unlike previous ones using a single loss for all parameters, effectively utilizes meta-learning's regularization to prevent overfitting.
MTA\cite{meng2023meta} is mainly conducting task augmentations by generating new labeled samples through retrieving highly relevant motifs from a pre-defined motif vocabulary as an external memory.
To utilize many-to-many correlations of molecules and properties, Zhuang et al.\cite{zhuang2023graph} construct a Molecule-Property relation Graph(MPG), then reformulate an episode in meta-learning as a subgraph of the MPG, and then schedule the subgraph sampling process with a contrastive loss function, which considers the consistency and discrimination of subgraphs.
Guo et al.\cite{guo2021few} developed a model where the importance of different property prediction tasks in few-shot learning is gauged using a self-attentive task weight, calculated by averaging molecular embeddings from each task's query set, to represent task significance.
Wang et al.\cite{wang2021property} propose a property-aware embedding function for context-based molecular adaptation and an adaptive relation graph module for molecular relation and embedding refinement, and then employ selective meta-learning strategy for task-specific parameter updates, effectively harmonizing shared knowledge and unique aspects in property prediction tasks.
Yao et al.\cite{yao2022chemical} picked out some molecules sharing common properties and use multiple property-aware graph neural networks to extract molecular representation, then employ the Spearman’s correlation to built property-aware matrix. In the few-shot MPP task, the meta-learning strategy is adopted to learn common prediction knowledge from the meta-training categories.\\
\indent In conclusion, transfer learning has gained popularity for its ability to enhance model generalization in scenarios with limited labeled data. This method is particularly effective in exploiting the relationships between molecules and properties, identifying shared information such as the role of molecular substructures across different tasks, which is crucial for developing more informed and accurate predictive models.
\\
\subsubsection{Multi-task Learning}
\noindent
Multi-task learning is a machine learning approach where a model is trained on multiple related tasks simultaneously, rather than training on each task independently. This strategy leverages the commonalities and differences across tasks, allowing the model to learn more generalizable features. \\
\indent Ma et al.\cite{ma2020improving} establishing a multi-label supervised model on a combined dataset with missing labels. the input to prediction network is a data matrix with multiple property label information, which can be an original dataset collected from specialized experiments.
Tan et al.\cite{tan2021multitask} constructed our multitask models by stacking a base regressor and classifier, enabling multitarget predictions through an additional training stage on the expanded molecular feature space.
Biswas et al.\cite{biswas2023predicting} employed a multitask training method for a single model to predict critical properties and acentric factors, while also adjusting target weights in the loss function to correct data imbalance.\\
\indent These works we've reviewed show that multi-task learning is highly effective in MPP, as it capitalizes on the interrelation of various molecular properties. 
This enhances a model's capacity to simultaneously predict multiple properties, which is a particularly valuable trait when dealing with the challenge of limited labeled data. 
\section{Evaluation and Benchmark}
\noindent
In evaluating the performance of models in Molecular Property Prediction (MPP), it is crucial to consider a variety of benchmarks, each offering distinct datasets and posing unique challenges. 
Key benchmarks include MoleculeNet\cite{wu2018moleculenet}, ADMETlab\cite{dong2018admetlab}, MoleculeACE\cite{van2022exposing}, DrugOOD\cite{ji2023drugood}, MD17\cite{chmiela2017machine}, TUDataset\cite{morris2020tudataset} (comprising MUTAG, PTC, NCI, PROTEINS, D\&D, and ENZYMES), and PCQM4Mv2\cite{hu2020open}; their details are shown in Table \ref{table:datasets}.
MoleculeNet, our primary focus, offers a diverse collection of datasets in quantum mechanics, physical chemistry, biophysics, and physiology, crucial for multifaceted molecular property predictions. ADMETlab, is vital for assessing drug safety and efficacy, providing data on ADMET properties. MoleculeACE focuses on QSAR modeling challenges, notably activity cliffs. DrugOOD, based on ChEMBL, emphasizes out-of-distribution generalization in AI-aided drug discovery. MD17 is essential for validating models in computational chemistry with its molecular dynamics trajectories. TUDataset includes varied datasets like DD, ENZYMES, PROTEINS, and MUTAG, each presenting unique graph-based bioinformatics challenges. Lastly, PCQM4Mv2 from the Open Graph Benchmark offers large-scale quantum mechanical property prediction challenges for graph neural network models. Among these, MoleculeNet stands out due to its comprehensive coverage and wide usage, making it an exemplary benchmark for our evaluation.\\
\indent MoleculeNet, a frequently used benchmark in MPP, offers a diverse range of datasets categorized into four groups: Quantum Mechanics, Physical Chemistry, Biophysics, and Physiology. Each group provides specialized datasets to assess different aspects of molecular properties:
Quantum Mechanics: Datasets in this group are centered around electronic properties derived from quantum mechanical calculations.
Physical Chemistry: These datasets focus on physical and chemical properties of molecules, including solubility and lipophilicity.
Biophysics: This category includes datasets related to biological interactions and processes, such as protein-ligand binding affinities.
Physiology: Datasets here pertain to organism-level effects, like toxicity and drug efficacy.
Evaluating models across these diverse datasets from MoleculeNet allows for a comprehensive assessment of their predictive capabilities in various aspects of MPP.\\
\indent Consistent with prior studies, we adopt the area under the receiver operating characteristic curve (ROC-AUC) as the evaluation metric for classification datasets, which is a widely used metric for assessing the performance of binary classification tasks.
For the regression datasets, we utilize root-mean-squared error (RMSE) as the evaluation metric. 
\begin{table*}[ht]
\centering
\caption{
\textbf{Overview of Datasets for Molecular Property Prediction.} This table encapsulates key benchmarks, highlighting their scale, scope, and specific applications in the fields of molecular modeling, drug discovery, and computational chemistry. "Num. of Mol." means the number of molecules in the corresponding Benchmark.\\
}
\label{table:datasets}
\begin{tabular}{p{2.5cm}|p{8cm}|p{1.8cm}|p{3cm}}
\toprule
\hline
\textbf{Benchmark Name} & \textbf{Description} & \textbf{Num. of Mol.} & \textbf{Application/Challenge} \\ \hline
MoleculeNet\cite{wu2018moleculenet} & A diverse collection of datasets across quantum mechanics, physical chemistry, and biophysical properties, pivotal for various molecular property predictions. & 785,951 & Multifaceted challenges in molecular property predictions \\ \hline
ADMETlab\cite{dong2018admetlab} & Provides extensive data on ADMET properties crucial for drug safety and efficacy assessments, enhancing drug development processes. & 94,387 & Drug development and safety evaluation \\ \hline
MoleculeACE\cite{van2022exposing} & Focused on QSAR modeling challenges, especially activity cliffs where minor structural changes cause significant bioactivity variations, testing the robustness of ML models. & 48,707 & Model accuracy in subtle molecular variations \\ \hline
DrugOOD\cite{ji2023drugood} & Based on ChEMBL, it emphasizes out-of-distribution (OOD) generalization, crucial for advancing AI in drug discovery under limited and varied data scenarios. & 930,314 & OOD generalization in AI-aided drug discovery \\ \hline
MD17\cite{chmiela2017machine} & Contains molecular dynamics trajectories, essential for developing and validating models in computational chemistry and molecular simulations. & 3,817,604 & Molecular dynamics model development and validation \\ \hline
TUDataset\cite{morris2020tudataset} (MUTAG, PTC, NCI, PROTEINS, D\&D, ENZYMES) & Includes datasets like DD, ENZYMES, PROTEINS, and MUTAG, each offering unique bioinformatics challenges in graph-based analysis, such as protein structure and enzyme function classification. & ---- & Bioinformatics applications in graph-based learning \\ \hline
PCQM4Mv2\cite{hu2020open} & A dataset from the Open Graph Benchmark, providing large-scale quantum mechanical property prediction challenges for graph neural network models. & 3,746,619 & Quantum mechanical property prediction in molecular systems \\ 
\hline
\bottomrule
\end{tabular}
\end{table*}
It's important to note that many studies in this field adopt either random or scaffold splits for dividing their datasets, though not uniformly. 
A random split involves randomly dividing the dataset into training, validation, and test sets, regardless of molecular structures. 
On the other hand, a scaffold split organizes molecules based on their core chemical scaffolds, ensuring that the model is tested on chemically distinct molecules from those it was trained on, providing a more stringent test of its generalization ability. 
The choice between these splitting methods can significantly affect the outcomes and interpretations of model performance evaluations.
\\
\section{Discussion}
\noindent
\subsection{Domain Knowledge Integration}
\begin{figure*}[!h]
    \centering
    \includegraphics[width=0.65\linewidth]{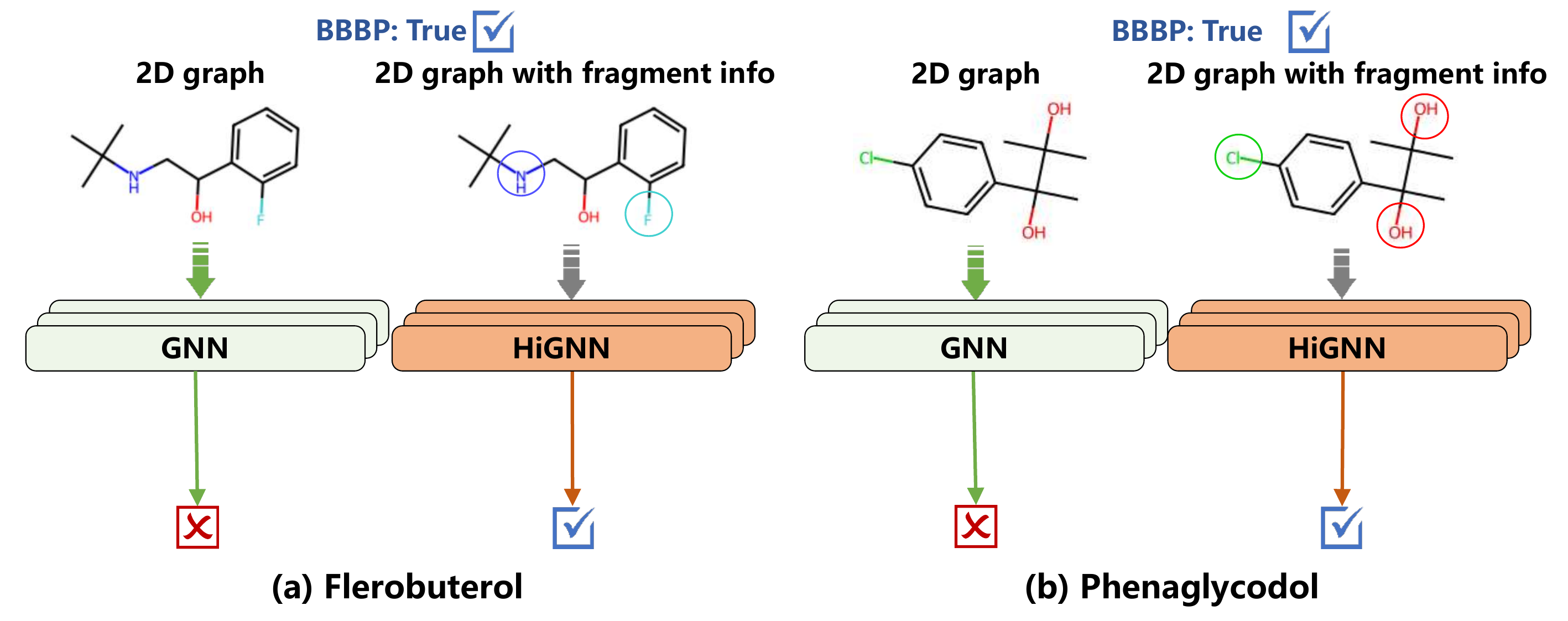}
    \caption{
    \textbf{Case study of molecular fragment information enhance for the BBBP Task.}
    This figure compares the predictions of HiGNN\cite{zhu2022hignn} for two molecules, Flerobuterol and Phenaglycodol, with and without the use of BRICS-derived molecular substructure information. On the left, Flerobuterol's molecular structure without substructure information leads to an incorrect BBBP prediction, while on the right, incorporating key fragments like the fluorine atom and tertiary amine yields an accurate prediction, highlighting these features' role in CNS activity and BBB penetration. Similarly, for Phenaglycodol, BRICS fragmentation reveals critical hydrophilic and lipophilic components, resulting in a correct BBBP prediction, demonstrating the model's improved predictive capability when domain knowledge is applied.
    }
    \label{fig:CS2}
\end{figure*}

\noindent
\begin{table*}[ht]
\centering
\caption{\textbf{Comparison of DL methods for MPP classification tasks with substructure domain knowledge in MoleculeNet}. This table contrasts various models, focusing on classification (ROC-AUC \%) tasks. Each model is evaluated with and without substructure information, as indicated by original and ablation study rows. The '-' symbol marks the absence of data for some datasets, while 'avg. imp.' shows the average performance improvement due to substructure information integration.\\
}
\resizebox{0.85\textwidth}{!}{%
\begin{tabular}{@{}c|c|cccccccc@{}}
\toprule
\hline
\multirow{2}{*}{\textbf{Model}} & \multirow{2}{*}{\textbf{Splitting}} & \multicolumn{7}{c}{Classification (ROC-AUC (\%) higher is better ↑)} & \multirow{2}{*}{\textbf{avg. imp.}}\\
& & \textbf{BBBP} & \textbf{Tox21} & \textbf{ToxCast} & \textbf{SIDER} & \textbf{ClinTox} & \textbf{BACE} & \textbf{HIV} &
\\
\hline
\multirow{2}{*}{\makecell{MoLGNN \cite{shen2020molgnn} \\ (MoLGNN, GINVAE only)}} & \multirow{2}{*}{random} & 88.9 & $-$ & $-$ & 63.6 & 94.2 & 87.4 & 78.0& \multirow{2}{*}{1.09\%}  \\
& & 89.2 & $-$ & $-$ & 61.7 & 93.7 & 87.1 & 76.3 &  \\
\hline
\multirow{2}{*}{\makecell{HiGNN \cite{zhu2022hignn} \\ (HiGNN, w/o HI)}}  & \multirow{2}{*}{random} & 93.2 & 85.6 & $-$ & 65.1 & 93.0 & 89.0 & $-$ & \multirow{2}{*}{0.25\%}  \\
& & 93.0 & 85.2 & $-$ & 65.4 & 92.6 & 88.7 & $-$ &  \\
\hline
\multirow{2}{*}{\makecell{MISU \cite{benjamin2022graph} \\ (MISU, w/o JTVAE)}} & \multirow{2}{*}{scaffold} & 66.7 & 76.3 & 62.8 & 59.7 & 78.0 & 70.5 & 
 $-$ & \multirow{2}{*}{1.93\%} \\
& & 65.9 & 76.2 & 62.3 & 58.4 & 76.1 & 67.1 & $-$ &  \\
\hline
\multirow{2}{*}{\makecell{CAFE \cite{xie2023self} \\ (CAFE-MPP, Only Graphormer)}} & \multirow{2}{*}{random} & 96.5 & 80.5 & $-$ & 65.8 & 98.2 & 93.9 & $-$ & \multirow{2}{*}{3.93\%} \\
& & 93.6 & 79.3 & $-$ & 61.8 & 94.3 & 89.1 & $-$ &  \\
\hline
\multirow{2}{*}{\makecell{iMolCLR \cite{wang2022improving} \\ (iMolCLR, MolCLR)}} & \multirow{2}{*}{scaffold} & 76.4 & 79.9 & 73.6 & 69.9 & 95.4 & 88.5 & 80.8 & \multirow{2}{*}{1.39\%} \\
& & 73.6 & 79.8 & 72.7 & 68.0 & 93.2 & 89.0 & 80.6 &  \\
\hline
\bottomrule
\end{tabular}
}
\label{tab: domain-knowledge-comparison-class}
\end{table*}
\begin{table*}[ht]
\centering
\caption{\textbf{Comparison of DL methods for MPP regression tasks with substructure domain knowledge in MoleculeNet}. This table contrasts various models, focusing on regression (RMSE) tasks.
Each model is evaluated with and without substructure
information, as indicated by original and ablation study rows. The ’-’ symbol marks the absence of data for some datasets, while ’avg. imp.’ shows the average performance improvement due to substructure information integration.\\
}
\resizebox{0.7\textwidth}{!}{%
\begin{tabular}{@{}c|c|cccccc@{}}
\toprule
\hline
\multirow{2}{*}{\textbf{Model}} & \multirow{2}{*}{\textbf{Splitting}} & \multicolumn{5}{c}{Regression (RMSE, lower is better ↓)} & \multirow{2}{*}{\textbf{avg. imp.}}\\
& & \textbf{ESOL} & \textbf{FreeSolv} & \textbf{Lipo} & \textbf{QM7} & \textbf{QM8} &
\\
\hline
\multirow{2}{*}{\makecell{HiGNN \cite{zhu2022hignn} \\ (HiGNN, w/o HI)}} & \multirow{2}{*}{random} & 0.532 & 0.915 & 0.549 & $-$ & $-$ & \multirow{2}{*}{2.78\%}   \\
& & 0.536 & 0.941 & 0.575 & $-$ & $-$ &  \\
\hline
\multirow{2}{*}{\makecell{CAFE \cite{xie2023self} \\ (CAFE-MPP, Only Graphormer)}} & \multirow{2}{*}{random} & 0.687 & 1.276 & 0.684 & 43.75 & 0.0141 & \multirow{2}{*}{1.37\%} \\
& & 0.782 & 1.303 & 0.718 & 40.69 & 0.0138 &  \\
\hline
\multirow{2}{*}{\makecell{iMolCLR \cite{wang2022improving} \\ (iMolCLR, MolCLR)}} & \multirow{2}{*}{scaffold} & 1.130 & 2.090 & 0.640 & 66.30 & 0.0170 & \multirow{2}{*}{7.78\%}  \\
& & 1.110 & 2.200 & 0.650 & 87.2 & 0.0174 &  \\
\hline
\bottomrule
\end{tabular}
}
\label{tab: domain-knowledge-comparison-regr}
\end{table*}
This part aims to analyze the contribution of domain knowledge for MPP, as the model input.
It is divided into 3 part: atom-bond property, molecular structure, and molecular property relation.\\
\indent As more research utilizes atom and bond properties, the efficiency of MPP has improved. However, it raises the question: does integrating additional atom and bond properties into the model input necessarily lead to higher model performance?
Wojtuch et al.\cite{wojtuch2023extended} analyzed the impact of atomic features in graph convolutional neural networks, comparing twelve hand-crafted and four literature-based feature combinations. 
Findings indicate that feature importance is task-specific and linked to their prevalence in the dataset. 
Reducing less frequent or redundant features, such as formal charges or aromaticity, improves performance. 
These insights also apply to advanced models like Graph Transformers, though optimal feature selection varies by model.\\
\indent Increasingly, molecular structure information is being incorporated into MPP, with several studies leveraging it to derive coarse-grained molecular insights from hierarchical graphs. 
Recent method like MoLGNN\cite{shen2020molgnn}, HiGNN\cite{zhu2022hignn}, MISU\cite{benjamin2022graph}, CAFE\cite{xie2023self}, and iMolCLR\cite{wang2022improving} have used molecular substructure knowledge, such as BRICS or functional groups, to construct hierarchical graphs treating fragments as nodes. 
These methods have shown improved results over those not using substructure information.
As Table \ref{tab: domain-knowledge-comparison-class} and Table \ref{tab: domain-knowledge-comparison-regr} demonstrate, the ablation studies reveal a notable enhancement in methodology efficacy when fragment or functional group information is integrated. 
Specifically, we observe a 3.98\% improvement in regression tasks, measured using RMSE, and a 1.72\% improvement in classification tasks, measured using ROC-AUC. These results confirm the significant impact of incorporating substructure domain knowledge into these deep learning models.
We present two compelling case studies that illustrate the impact of molecular substructure information obtained via the BRICS methodology. The first example involves the molecule 'CC(C)(C)NCC(O)c1ccccc1F'. When employing BRICS fragmentation, the model identifies the fluorine atom and the tertiary amine as critical features. These fragments are known to significantly affect CNS activity due to their lipophilicity, which is a crucial determinant for blood-brain barrier (BBB) penetration. The second case focuses on CC(C)(O)C(C)(O)c1ccc(Cl)cc1, where BRICS fragmentation reveals the delicate balance between hydrophilic hydroxyl groups and lipophilic chlorinated benzene components. This balance plays a pivotal role in the molecule's ability to penetrate the BBB. Both examples, depicted in Figure \ref{fig:CS2}, showcase enhanced performance of HiGNN\cite{zhu2022hignn} when integrating substructure information, confirming the model's superior ability to predict BBB penetration by capturing intricate substructure information.

\indent Identifying a fundamental set of properties for molecular prediction tasks is crucial for future research. 
Many studies, including multi-task learning methods, have shown that fundamental molecular properties can enhance other prediction tasks. 
For instance, Sun et al.\cite{sun2022pemp} improved the training of chemical and physiological property predictors by incorporating related physics property prediction tasks. 
Additionally, Biswas et al.\cite{biswas2023predicting} demonstrated the significance of critical properties and acentric factors, along with four phase change properties as auxiliary targets.\\
\indent However, the integration of domain knowledge into molecular property prediction models is not without challenges. 
Firstly, there is still a lot of domain knowledge that is not digitized or gathered, even with the advances in tools like RDKit. It is possible to overlook important subtleties when converting intricate, frequently implicit expert information into an electronic format that is easy to use. 
Secondly, this integration process can introduce biases due to subjective interpretations by domain experts, potentially skewing model outcomes and impacting scalability and adaptability to new molecular data types. 
Lastly, the requirement to customize deep learning architectures to incorporate such knowledge significantly increases complexity and computational costs, complicating model development and training.\\
\indent In conclusion, while domain knowledge integration is beneficial, it necessitates a careful and balanced approach. It is crucial to maintain the flexibility, scalability, and objectivity of models.
These challenges highlight the need for ongoing efforts to capture and digitize comprehensive domain knowledge, maintaining a critical balance between accuracy and the practical application of these predictive models.

\subsection{Multi-modal Data Utilization}
\noindent
\begin{figure*}[!h]
    \centering
    \includegraphics[width=1.0\linewidth]{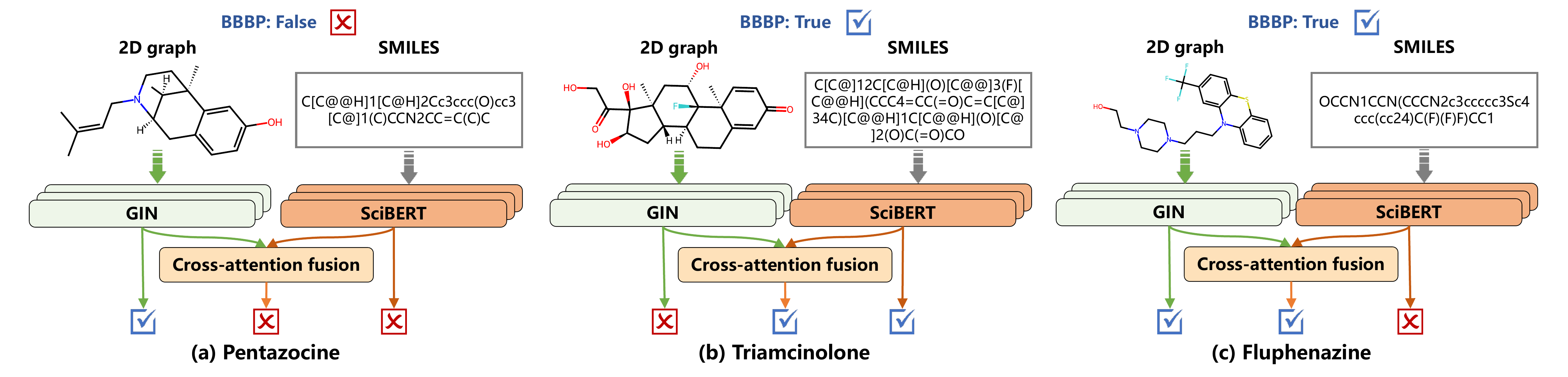}
    \caption{\textbf{Case studies of multi-modal fusion for the BBBP Task.}
    Some molecules, such as pentazocine and triamcinolone, may give incorrect predictions when based solely on 2D graph data.
    However, integrating SMILES information can correct these results.
    On the other hand, predictions based on 2D graphs for molecules like fluphenazine are accurate, whereas SMILES predictions are not.
    Nevertheless, combining both modalities does not compromise the overall accuracy of the final assessment.
    This underscores the integration of 2D graph modality data with SMILES information, which can enhance the model's ability to correct erroneous predictions and safeguard against potential interferences.}
    \label{fig:STCS}
\end{figure*}
\begin{table*}[ht]
\centering
\caption{\textbf{Comparison of multi-modal learning methods in MoleculeNet.}
This table contrasts various models, focusing on classification (ROC-AUC \%) tasks.
The `Methods' column specifies the learning strategy: `T' denotes the use of textual data, `S' denotes the use of SMILES data, `2d' and `3d' refer to the use of 2D and 3D molecular graphs, `CL' indicates contrastive learning, and `CA' stands for cross-attention fusion.
The columns—SMILES data, graph2d, and graph3d—with the \checkmark highlight the types of input data utilized by the models, with the possibility of multiple selections.
The best results are \textbf{emboldened}, and the second-best results are highlighted in \textcolor{red}{red}.\\}
\resizebox{\textwidth}{!}{%
\begin{tabular}{@{}c|c|c|ccc|ccccccc@{}}
\toprule
\hline
\multirow{2}{*}{\textbf{Model}} & 
\multirow{2}{*}{\textbf{Methods}} & 
\multirow{2}{*}{\textbf{Index}} &
\multicolumn{3}{c|}{Type of input data}& 
\multicolumn{6}{c}{Classification (ROC-AUC \% higher is better ↑)}&
\multirow{2}{*}{\textbf{Average}} \\
& & &\textbf{S} & \textbf{2d} & \textbf{3d}& 
\textbf{BBBP} & \textbf{Tox21} & \textbf{ToxCast} & \textbf{SIDER} & \textbf{ClinTox} & \textbf{BACE} &
\\
\hline
KV-PLM \cite{zeng2022deep} & \textbackslash & \textbf{0} &\checkmark & & & 72.0 & 70.0 & 55.0 & 59.8 & 89.2 & 78.5 & 70.8 \\
\hline
GIN \cite{xu2018powerful} & \textbackslash & \textbf{1} & & \checkmark & & 65.4 & 74.9 & 61.6 & 58.0 & 58.8 & 72.6 & 65.2 \\
\hline
\multirow{2}{*}{\makecell{GraphMVP \cite{liu2021pre} \\ (GIN, SchNet)}} & CL(2d, 3d) & \textbf{2} & & \checkmark &  & 68.5 & 74.5 & 62.7 & 62.3 & 79.0 & 76.8 & 71.7 \\
\cline{2-13}
& CL(2d, 3d), CL(2d) & \textbf{3} & & \checkmark & &  72.4 & 74.4 & 63.1 & 63.9 & 77.5 & 81.2 & 72.1 \\
\hline
MoMu-S \cite{su2022molecular} & \multirow{4}{*}{CL(S, 2d)} & \textbf{4} & & \checkmark & & 70.5 & 75.6 & 63.4 & 60.5 & 79.9 & 76.7 & 71.1 \\
\cline{1-1}
\cline{3-13}
MoMu-K \cite{su2022molecular} & & \textbf{5} &  & \checkmark & & 70.1 & 75.6 & 63.0 & 60.4 & 77.4 & 77.1 & 70.6 \\
\cline{1-1}
\cline{3-13}
\multirow{2}{*}{\makecell{MoleculeSTM \cite{liu2023multi} \\ (MegaMolBART, GIN)}} & & \textbf{6}& \checkmark & & & 70.8 & 75.7 & 65.2 & 63.7 & 86.6 & 82.0 & 74.0 \\
& & \textbf{7}& & \checkmark & & 70.0 & 76.9 & 65.1 & 61.0 & \textbf{92.5} & 80.8 & 74.4 \\
\hline
\multirow{2}{*}{\makecell{MolFM \cite{luo2023molfm}\\ (KV-PLM, GIN)}} & \multirow{5}{*}{CL(S, 2d), CA} & \textbf{8}& & \checkmark & & 72.2 & 76.6 & 64.2 & 63.2 & 78.6 & 82.6 & 72.9 \\
& & \textbf{9}& \checkmark & \checkmark & & 72.9 & \textcolor{red}{77.2} & 64.4 & \textcolor{red}{64.2} & 79.7 & \textcolor{red}{83.9} & 73.7 \\
\cline{1-1}
\cline{3-13}
\multirow{3}{*}{\makecell{GIT-Mol \cite{liu2024git}\\ (SciBERT, MoMu-S)}}& & \textbf{10} & \checkmark & & & 71.9 & 73.9 & 62.1 & 60.1 & 83.5 & 68.4 & 70.0  \\
& & \textbf{11}& & \checkmark & & 71.1 & 75.4 & 65.3 & 58.2 & 78.9 & 65.8 & 69.1 \\
& & \textbf{12}& \checkmark & \checkmark & & \textcolor{red}{73.9} & 75.9 & \textcolor{red}{66.8} & 63.4 & 88.3 & 81.1 & \textcolor{red}{74.9}  \\
\hline
MolLM \cite{tang2023mollm} & CL(T, 2d), CL(T, 3d)  & \textbf{13} & &\checkmark & \checkmark & \textbf{75.7} & \textbf{80.0} & \textbf{68.2} & \textbf{71.0} & \textcolor{red}{91.1} & \textbf{84.1} & \textbf{78.4}  \\
\hline
\bottomrule
\end{tabular}
}
\label{tab: multi-modal-comparison}
\end{table*}
This part aims to analyze the contribution of different modalities in multi-modal models for MPP.
It delves into understanding how individual and combined modalities affect prediction performance.
Data from various studies are collated and analyzed, emphasizing the contribution of different modalities to MPP tasks, with models in Table \ref{tab: multi-modal-comparison}.
Apart from ClinTox, there is uniformity in the predictive prowess displayed by all models across a spectrum of tasks.
Nonetheless, ClinTox predictions are prone to biases in Transformer-based models due to data distribution peculiarities, which result in polarized predictions.
Graph-based models like GraphMVP, MoMu, and GIT-Mol(2d) demonstrate a reduction in such bias, albeit with compromised performance in ClinTox.

From an input modality perspective, taking the BBBP task of 2d graph and SMILES information fusion as an example from GIT-Mol\cite{liu2024git}, the size of the test dataset by scaffold split is 204.
We select study cases in which the results are superior to the baseline (SciBERT) after modality fusion, as shown in Figure \ref{fig:STCS}.
This illustration reveals the beneficial role of SMILES modality data in augmenting graph2d data,  whereby the integrated representation vectors can rectify erroneous predictions to a certain extent.
Conversely, accurate predictions from graph2d, when paired with incorrect SMILES predictions,  can also prevent potential mistakes,  showcasing the complementary strengths of integrating diverse modalities in enhancing predictive accuracy.

In examining pre-training strategies, contrastive learning clearly demonstrates significant benefits. 
In examining pre-training strategies, contrastive learning clearly demonstrates significant benefits. 
However, the integration of cross-attention might inadvertently reduce the impact of the singular modalities. Nonetheless, the strategic implementation of cross-attention promotes an effective fusion of SMILES and graph2d, resulting in combined vectors that outperform the individual modalities.

As shown in Table \ref{tab: multi-modal-comparison}, the methods involving modality alignment, which utilize contrastive learning between SMILES and 2D graphs, improve the model's performance from 65.2\% (index [1]) to 72.0\% (average of indexes [4, 5, 7]).
Furthermore, the methods integrating cross-attention mechanisms for modality fusion further enhance the model's performance to 74.3\% (average of indexes [9, 12]).
GraphMVP, using contrastive learning between 2D and 3D graphs, elevates the performance from 65.2\% to 71.7\% (indexes [1, 2]).
The MolLM achieves the optimal performance of 78.4\% (index [13]) through contrastive learning and the fusion of 2D and 3D graphs.

This reveals that multi-modal learning based on 2D graphs offers a performance increase of 6.5\% (71.7\% - 65.2\%) to 6.8\% (72.0\% - 65.2\%) over single-modality learning, with the attention fusion mechanism providing an additional 2.3\% (74.3\% - 72.0\%) to 6.7\% (78.4\% - 71.7\%) boost.

In the realm of molecular property prediction, the application of multi-modality methods introduces significant challenges and limitations. Key among these is the substantial increase in computational resource consumption required for processing complex multi-modal data. This issue is particularly evident when generating detailed 3D representations from standard molecular formats like SMILES, which demands extensive resources. Concurrently, these methods often contend with processing redundant information, as a result of overlapping content among various modalities, such as SMILES, 2D graphs, and 3D structures. This overlap leads to inefficiencies due to the repeated processing of identical molecular characteristics. Additionally, integrating different data forms into a cohesive model adds another layer of complexity, necessitating a strategic approach for effective data combination to enhance predictive accuracy.\\
\indent Despite these challenges, the conclusion drawn from our findings is clear: by anchoring on 2D graphs and enriching them with 1D SMILES or 3D graph information, multi-modal learning has achieved a significant ROC-AUC uplift of 9.1\% to 13.2\% compared to single-modality models (results from indexes [1, 9, 12, 13] in Table \ref{tab: multi-modal-comparison}).
These results underscore the substantial advantages and vast potential of modality fusion techniques in providing more holistic and comprehensive insights into molecular structures, thus enhancing the overall predictive accuracy in molecular property prediction.
\section{Conclusion}
\noindent
In this paper, we discuss the significant role of multi-modal data and domain knowledge in enhancing molecular property prediction through DL methods. 
We explored various molecular modalities and domain knowledge, crucial in understanding molecular complexities. 
Our review of recent encoder architectures and training strategies highlighted how integrating domain knowledge and multi-modal data advances these models. 
By benchmarking prominent works, we provided a comparative analysis of their effectiveness. 
Ultimately, our discussion revealed the profound impact of domain knowledge and multi-modal data in DL approaches, marking a transformative advancement in drug discovery and computational molecular analysis.
\bibliographystyle{unsrt} 
\bibliography{arxiv.bib}

\begin{thebibliography}{100}

\bibitem{shen2019molecular}
Jie Shen and Christos~A Nicolaou.
\newblock Molecular property prediction: recent trends in the era of artificial
  intelligence.
\newblock {\em Drug Discovery Today: Technologies}, 32:29--36, 2019.

\bibitem{li2022deep}
Zhen Li, Mingjian Jiang, Shuang Wang, and Shugang Zhang.
\newblock Deep learning methods for molecular representation and property
  prediction.
\newblock {\em Drug Discovery Today}, page 103373, 2022.

\bibitem{ma2020improving}
Hehuan Ma, Chaochao Yan, Yuzhi Guo, Sheng Wang, Yuhong Wang, Hongmao Sun, and
  Junzhou Huang.
\newblock Improving molecular property prediction on limited data with deep
  multi-label learning.
\newblock In {\em 2020 IEEE International Conference on Bioinformatics and
  Biomedicine (BIBM)}, pages 2779--2784. IEEE, 2020.

\bibitem{lin2020novel}
Xuan Lin, Zhe Quan, Zhi-Jie Wang, Huang Huang, and Xiangxiang Zeng.
\newblock A novel molecular representation with bigru neural networks for
  learning atom.
\newblock {\em Briefings in bioinformatics}, 21(6):2099--2111, 2020.

\bibitem{lv2021mol2context}
Qiujie Lv, Guanxing Chen, Lu~Zhao, Weihe Zhong, and Calvin Yu-Chian~Chen.
\newblock Mol2context-vec: learning molecular representation from context
  awareness for drug discovery.
\newblock {\em Briefings in Bioinformatics}, 22(6):bbab317, 2021.

\bibitem{han2023himgnn}
Shen Han, Haitao Fu, Yuyang Wu, Ganglan Zhao, Zhenyu Song, Feng Huang, Zhongfei
  Zhang, Shichao Liu, and Wen Zhang.
\newblock Himgnn: a novel hierarchical molecular graph representation learning
  framework for property prediction.
\newblock {\em Briefings in Bioinformatics}, 24(5):bbad305, 2023.

\bibitem{bouritsas2022improving}
Giorgos Bouritsas, Fabrizio Frasca, Stefanos Zafeiriou, and Michael~M
  Bronstein.
\newblock Improving graph neural network expressivity via subgraph isomorphism
  counting.
\newblock {\em IEEE Transactions on Pattern Analysis and Machine Intelligence},
  45(1):657--668, 2022.

\bibitem{song2020communicative}
Ying Song, Shuangjia Zheng, Zhangming Niu, Zhang-Hua Fu, Yutong Lu, and Yuedong
  Yang.
\newblock Communicative representation learning on attributed molecular graphs.
\newblock In {\em IJCAI}, volume 2020, pages 2831--2838, 2020.

\bibitem{li2022kpgt}
Han Li, Dan Zhao, and Jianyang Zeng.
\newblock Kpgt: knowledge-guided pre-training of graph transformer for
  molecular property prediction.
\newblock In {\em Proceedings of the 28th ACM SIGKDD Conference on Knowledge
  Discovery and Data Mining}, pages 857--867, 2022.

\bibitem{xiong2019pushing}
Zhaoping Xiong, Dingyan Wang, Xiaohong Liu, Feisheng Zhong, Xiaozhe Wan, Xutong
  Li, Zhaojun Li, Xiaomin Luo, Kaixian Chen, Hualiang Jiang, et~al.
\newblock Pushing the boundaries of molecular representation for drug discovery
  with the graph attention mechanism.
\newblock {\em Journal of medicinal chemistry}, 63(16):8749--8760, 2019.

\bibitem{rong2020self}
Yu~Rong, Yatao Bian, Tingyang Xu, Weiyang Xie, Ying Wei, Wenbing Huang, and
  Junzhou Huang.
\newblock Self-supervised graph transformer on large-scale molecular data.
\newblock {\em Advances in Neural Information Processing Systems},
  33:12559--12571, 2020.

\bibitem{ross2022large}
Jerret Ross, Brian Belgodere, Vijil Chenthamarakshan, Inkit Padhi, Youssef
  Mroueh, and Payel Das.
\newblock Large-scale chemical language representations capture molecular
  structure and properties.
\newblock {\em Nature Machine Intelligence}, 4(12):1256--1264, 2022.

\bibitem{yin2023lgi}
Shuo Yin and Guoqiang Zhong.
\newblock Lgi-gt: graph transformers with local and global operators
  interleaving.
\newblock In {\em Proceedings of the Thirty-Second International Joint
  Conference on Artificial Intelligence, IJCAI-23}, pages 4504--4512, 2023.

\bibitem{luo2022one}
Shengjie Luo, Tianlang Chen, Yixian Xu, Shuxin Zheng, Tie-Yan Liu, Liwei Wang,
  and Di~He.
\newblock One transformer can understand both 2d \& 3d molecular data.
\newblock {\em arXiv preprint arXiv:2210.01765}, 2022.

\bibitem{zeng2022accurate}
Xiangxiang Zeng, Hongxin Xiang, Linhui Yu, Jianmin Wang, Kenli Li, Ruth
  Nussinov, and Feixiong Cheng.
\newblock Accurate prediction of molecular properties and drug targets using a
  self-supervised image representation learning framework.
\newblock {\em Nature Machine Intelligence}, 4(11):1004--1016, 2022.

\bibitem{chen2021different}
Jen-Hao Chen and Yufeng~Jane Tseng.
\newblock Different molecular enumeration influences in deep learning: an
  example using aqueous solubility.
\newblock {\em Briefings in Bioinformatics}, 22(3):bbaa092, 2021.

\bibitem{liu2019multiresolution}
Shuai Liu, Jie Li, Kochise~C Bennett, Brad Ganoe, Tim Stauch, Martin
  Head-Gordon, Alexander Hexemer, Daniela Ushizima, and Teresa Head-Gordon.
\newblock Multiresolution 3d-densenet for chemical shift prediction in nmr
  crystallography.
\newblock {\em The journal of physical chemistry letters}, 10(16):4558--4565,
  2019.

\bibitem{liu2021pre}
Shengchao Liu, Hanchen Wang, Weiyang Liu, Joan Lasenby, Hongyu Guo, and Jian
  Tang.
\newblock Pre-training molecular graph representation with 3d geometry.
\newblock {\em arXiv preprint arXiv:2110.07728}, 2021.

\bibitem{li2022geomgcl}
Shuangli Li, Jingbo Zhou, Tong Xu, Dejing Dou, and Hui Xiong.
\newblock Geomgcl: Geometric graph contrastive learning for molecular property
  prediction.
\newblock In {\em Proceedings of the AAAI conference on artificial
  intelligence}, volume~36, pages 4541--4549, 2022.

\bibitem{zhu2022unified}
Jinhua Zhu, Yingce Xia, Lijun Wu, Shufang Xie, Tao Qin, Wengang Zhou, Houqiang
  Li, and Tie-Yan Liu.
\newblock Unified 2d and 3d pre-training of molecular representations.
\newblock In {\em Proceedings of the 28th ACM SIGKDD Conference on Knowledge
  Discovery and Data Mining}, pages 2626--2636, 2022.

\bibitem{guo2020graseq}
Zhichun Guo, Wenhao Yu, Chuxu Zhang, Meng Jiang, and Nitesh~V Chawla.
\newblock Graseq: graph and sequence fusion learning for molecular property
  prediction.
\newblock In {\em Proceedings of the 29th ACM international conference on
  information \& knowledge management}, pages 435--443, 2020.

\bibitem{wang2022molecular}
Yuyang Wang, Jianren Wang, Zhonglin Cao, and Amir Barati~Farimani.
\newblock Molecular contrastive learning of representations via graph neural
  networks.
\newblock {\em Nature Machine Intelligence}, 4(3):279--287, 2022.

\bibitem{fang2023knowledge}
Yin Fang, Qiang Zhang, Ningyu Zhang, Zhuo Chen, Xiang Zhuang, Xin Shao, Xiaohui
  Fan, and Huajun Chen.
\newblock Knowledge graph-enhanced molecular contrastive learning with
  functional prompt.
\newblock {\em Nature Machine Intelligence}, pages 1--12, 2023.

\bibitem{li2023knowledge}
Han Li, Ruotian Zhang, Yaosen Min, Dacheng Ma, Dan Zhao, and Jianyang Zeng.
\newblock A knowledge-guided pre-training framework for improving molecular
  representation learning.
\newblock {\em Nature Communications}, 14(1):7568, 2023.

\bibitem{hao2020asgn}
Zhongkai Hao, Chengqiang Lu, Zhenya Huang, Hao Wang, Zheyuan Hu, Qi~Liu, Enhong
  Chen, and Cheekong Lee.
\newblock Asgn: An active semi-supervised graph neural network for molecular
  property prediction.
\newblock In {\em Proceedings of the 26th ACM SIGKDD International Conference
  on Knowledge Discovery \& Data Mining}, pages 731--752, 2020.

\bibitem{sun2019infograph}
Fan-Yun Sun, Jordan Hoffmann, Vikas Verma, and Jian Tang.
\newblock Infograph: Unsupervised and semi-supervised graph-level
  representation learning via mutual information maximization.
\newblock {\em arXiv preprint arXiv:1908.01000}, 2019.

\bibitem{zhang2023dropconn}
Dan Zhang, Wenzheng Feng, Yuandong Wang, Zhongang Qi, Ying Shan, and Jie Tang.
\newblock Dropconn: Dropout connection based random gnns for molecular property
  prediction.
\newblock {\em IEEE Transactions on Knowledge and Data Engineering}, 2023.

\bibitem{sun2022pemp}
Yuancheng Sun, Yimeng Chen, Weizhi Ma, Wenhao Huang, Kang Liu, Zhiming Ma,
  Wei-Ying Ma, and Yanyan Lan.
\newblock Pemp: Leveraging physics properties to enhance molecular property
  prediction.
\newblock In {\em Proceedings of the 31st ACM International Conference on
  Information \& Knowledge Management}, pages 3505--3513, 2022.

\bibitem{chen2022meta}
Wenlin Chen, Austin Tripp, and Jos{\'e}~Miguel Hern{\'a}ndez-Lobato.
\newblock Meta-learning adaptive deep kernel gaussian processes for molecular
  property prediction.
\newblock In {\em The Eleventh International Conference on Learning
  Representations}, 2022.

\bibitem{zhuang2023graph}
Xiang Zhuang, Qiang Zhang, Bin Wu, Keyan Ding, Yin Fang, and Huajun Chen.
\newblock Graph sampling-based meta-learning for molecular property prediction.
\newblock {\em arXiv preprint arXiv:2306.16780}, 2023.

\bibitem{biswas2023predicting}
Sayandeep Biswas, Yunsie Chung, Josephine Ramirez, Haoyang Wu, and William~H
  Green.
\newblock Predicting critical properties and acentric factors of fluids using
  multitask machine learning.
\newblock {\em Journal of Chemical Information and Modeling},
  63(15):4574--4588, 2023.

\bibitem{tan2021multitask}
Zheng Tan, Yan Li, Weimei Shi, and Shiqing Yang.
\newblock A multitask approach to learn molecular properties.
\newblock {\em Journal of Chemical Information and Modeling}, 61(8):3824--3834,
  2021.

\bibitem{wu2018moleculenet}
Zhenqin Wu, Bharath Ramsundar, Evan~N Feinberg, Joseph Gomes, Caleb Geniesse,
  Aneesh~S Pappu, Karl Leswing, and Vijay Pande.
\newblock Moleculenet: a benchmark for molecular machine learning.
\newblock {\em Chemical science}, 9(2):513--530, 2018.

\bibitem{weininger1988smiles}
David Weininger.
\newblock Smiles, a chemical language and information system. 1. introduction
  to methodology and encoding rules.
\newblock {\em Journal of chemical information and computer sciences},
  28(1):31--36, 1988.

\bibitem{weininger1989smiles}
David Weininger, Arthur Weininger, and Joseph~L Weininger.
\newblock Smiles. 2. algorithm for generation of unique smiles notation.
\newblock {\em Journal of chemical information and computer sciences},
  29(2):97--101, 1989.

\bibitem{weininger1990smiles}
David Weininger.
\newblock Smiles. 3. depict. graphical depiction of chemical structures.
\newblock {\em Journal of chemical information and computer sciences},
  30(3):237--243, 1990.

\bibitem{rogers2010extended}
David Rogers and Mathew Hahn.
\newblock Extended-connectivity fingerprints.
\newblock {\em Journal of chemical information and modeling}, 50(5):742--754,
  2010.

\bibitem{durant2002reoptimization}
Joseph~L Durant, Burton~A Leland, Douglas~R Henry, and James~G Nourse.
\newblock Reoptimization of mdl keys for use in drug discovery.
\newblock {\em Journal of chemical information and computer sciences},
  42(6):1273--1280, 2002.

\bibitem{krenn2020self}
Mario Krenn, Florian H{\"a}se, AkshatKumar Nigam, Pascal Friederich, and Alan
  Aspuru-Guzik.
\newblock Self-referencing embedded strings (selfies): A 100\% robust molecular
  string representation.
\newblock {\em Machine Learning: Science and Technology}, 1(4):045024, 2020.

\bibitem{mcnaught1997compendium}
Alan~D McNaught, Andrew Wilkinson, et~al.
\newblock {\em Compendium of chemical terminology}, volume 1669.
\newblock Blackwell Science Oxford, 1997.

\bibitem{heller2015inchi}
Stephen~R Heller, Alan McNaught, Igor Pletnev, Stephen Stein, and Dmitrii
  Tchekhovskoi.
\newblock Inchi, the iupac international chemical identifier.
\newblock {\em Journal of cheminformatics}, 7(1):1--34, 2015.

\bibitem{landrum2013rdkit}
Greg Landrum et~al.
\newblock Rdkit: A software suite for cheminformatics, computational chemistry,
  and predictive modeling.
\newblock {\em Greg Landrum}, 8:31, 2013.

\bibitem{delano2002pymol}
Warren~L DeLano et~al.
\newblock Pymol: An open-source molecular graphics tool.
\newblock {\em CCP4 Newsl. Protein Crystallogr}, 40(1):82--92, 2002.

\bibitem{sunseri2020libmolgrid}
Jocelyn Sunseri and David~R Koes.
\newblock Libmolgrid: graphics processing unit accelerated molecular gridding
  for deep learning applications.
\newblock {\em Journal of chemical information and modeling}, 60(3):1079--1084,
  2020.

\bibitem{degen2008art}
J{\"o}rg Degen, Christof Wegscheid-Gerlach, Andrea Zaliani, and Matthias Rarey.
\newblock On the art of compiling and using'drug-like'chemical fragment spaces.
\newblock {\em ChemMedChem: Chemistry Enabling Drug Discovery},
  3(10):1503--1507, 2008.

\bibitem{lewell1998recap}
Xiao~Qing Lewell, Duncan~B Judd, Stephen~P Watson, and Michael~M Hann.
\newblock Recap retrosynthetic combinatorial analysis procedure: a powerful new
  technique for identifying privileged molecular fragments with useful
  applications in combinatorial chemistry.
\newblock {\em Journal of chemical information and computer sciences},
  38(3):511--522, 1998.

\bibitem{bemis1996properties}
Guy~W Bemis and Mark~A Murcko.
\newblock The properties of known drugs. 1. molecular frameworks.
\newblock {\em Journal of medicinal chemistry}, 39(15):2887--2893, 1996.

\bibitem{liu2017break}
Tairan Liu, Misagh Naderi, Chris Alvin, Supratik Mukhopadhyay, and Michal
  Brylinski.
\newblock Break down in order to build up: decomposing small molecules for
  fragment-based drug design with e molfrag.
\newblock {\em Journal of chemical information and modeling}, 57(4):627--631,
  2017.

\bibitem{kruger2020rdscaffoldnetwork}
Franziska Kruger, Nikolaus Stiefl, and Gregory~A Landrum.
\newblock rdscaffoldnetwork: the scaffold network implementation in rdkit.
\newblock {\em Journal of Chemical Information and Modeling}, 60(7):3331--3335,
  2020.

\bibitem{wu2021learning}
Cheng-Kun Wu, Xiao-Chen Zhang, Zhi-Jiang Yang, Ai-Ping Lu, Ting-Jun Hou, and
  Dong-Sheng Cao.
\newblock Learning to smiles: Ban-based strategies to improve latent
  representation learning from molecules.
\newblock {\em Briefings in Bioinformatics}, 22(6):bbab327, 2021.

\bibitem{yang2019analyzing}
Kevin Yang, Kyle Swanson, Wengong Jin, Connor Coley, Philipp Eiden, Hua Gao,
  Angel Guzman-Perez, Timothy Hopper, Brian Kelley, Miriam Mathea, et~al.
\newblock Analyzing learned molecular representations for property prediction.
\newblock {\em Journal of chemical information and modeling}, 59(8):3370--3388,
  2019.

\bibitem{ji2023metapath}
Ying Ji, Guojia Wan, Yibing Zhan, and Bo~Du.
\newblock Metapath-fused heterogeneous graph network for molecular property
  prediction.
\newblock {\em Information Sciences}, 629:155--168, 2023.

\bibitem{fang2022geometry}
Xiaomin Fang, Lihang Liu, Jieqiong Lei, Donglong He, Shanzhuo Zhang, Jingbo
  Zhou, Fan Wang, Hua Wu, and Haifeng Wang.
\newblock Geometry-enhanced molecular representation learning for property
  prediction.
\newblock {\em Nature Machine Intelligence}, 4(2):127--134, 2022.

\bibitem{zhou2023unimol}
Gengmo Zhou, Zhifeng Gao, Qiankun Ding, Hang Zheng, Hongteng Xu, Zhewei Wei,
  Linfeng Zhang, and Guolin Ke.
\newblock Uni-mol: A universal 3d molecular representation learning framework.
\newblock In {\em The Eleventh International Conference on Learning
  Representations}, 2023.

\bibitem{chithrananda2020chemberta}
Seyone Chithrananda, Gabriel Grand, and Bharath Ramsundar.
\newblock Chemberta: large-scale self-supervised pretraining for molecular
  property prediction.
\newblock {\em arXiv preprint arXiv:2010.09885}, 2020.

\bibitem{yuksel2023selformer}
Atakan Y{\"u}ksel, Erva Ulusoy, Atabey {\"U}nl{\"u}, and Tunca Do{\u{g}}an.
\newblock Selformer: Molecular representation learning via selfies language
  models.
\newblock {\em Machine Learning: Science and Technology}, 2023.

\bibitem{zhang2022abc}
Xiao-Chen Zhang, Jia-Cai Yi, Guo-Ping Yang, Cheng-Kun Wu, Ting-Jun Hou, and
  Dong-Sheng Cao.
\newblock Abc-net: a divide-and-conquer based deep learning architecture for
  smiles recognition from molecular images.
\newblock {\em Briefings in Bioinformatics}, 23(2):bbac033, 2022.

\bibitem{liu2023multi}
Shengchao Liu, Weili Nie, Chengpeng Wang, Jiarui Lu, Zhuoran Qiao, Ling Liu,
  Jian Tang, Chaowei Xiao, and Animashree Anandkumar.
\newblock Multi-modal molecule structure--text model for text-based retrieval
  and editing.
\newblock {\em Nature Machine Intelligence}, 5(12):1447--1457, 2023.

\bibitem{liu2015multi}
Pengfei Liu, Xipeng Qiu, Xinchi Chen, Shiyu Wu, and Xuan-Jing Huang.
\newblock Multi-timescale long short-term memory neural network for modelling
  sentences and documents.
\newblock In {\em Proceedings of the 2015 conference on empirical methods in
  natural language processing}, pages 2326--2335, 2015.

\bibitem{chung2015gated}
Junyoung Chung, Caglar Gulcehre, Kyunghyun Cho, and Yoshua Bengio.
\newblock Gated feedback recurrent neural networks.
\newblock In {\em International conference on machine learning}, pages
  2067--2075. PMLR, 2015.

\bibitem{nazarova2021dielectric}
Antonina~L Nazarova, Liqiu Yang, Kuang Liu, Ankit Mishra, Rajiv~K Kalia,
  Ken-ichi Nomura, Aiichiro Nakano, Priya Vashishta, and Pankaj Rajak.
\newblock Dielectric polymer property prediction using recurrent neural
  networks with optimizations.
\newblock {\em Journal of Chemical Information and Modeling}, 61(5):2175--2186,
  2021.

\bibitem{wang2019predictive}
Zihao Wang, Yang Su, Weifeng Shen, Saimeng Jin, James~H Clark, Jingzheng Ren,
  and Xiangping Zhang.
\newblock Predictive deep learning models for environmental properties: the
  direct calculation of octanol--water partition coefficients from molecular
  graphs.
\newblock {\em Green Chemistry}, 21(16):4555--4565, 2019.

\bibitem{withnall2020building}
Michael Withnall, Edvard Lindel{\"o}f, Ola Engkvist, and Hongming Chen.
\newblock Building attention and edge message passing neural networks for
  bioactivity and physical--chemical property prediction.
\newblock {\em Journal of cheminformatics}, 12(1):1--18, 2020.

\bibitem{li2021trimnet}
Pengyong Li, Yuquan Li, Chang-Yu Hsieh, Shengyu Zhang, Xianggen Liu, Huanxiang
  Liu, Sen Song, and Xiaojun Yao.
\newblock Trimnet: learning molecular representation from triplet messages for
  biomedicine.
\newblock {\em Briefings in Bioinformatics}, 22(4):bbaa266, 2021.

\bibitem{zhang2022coatgin}
Xuan Zhang, Cheng Chen, Zhaoxu Meng, Zhenghe Yang, Haitao Jiang, and Xuefeng
  Cui.
\newblock Coatgin: Marrying convolution and attention for graph-based molecule
  property prediction.
\newblock In {\em 2022 IEEE International Conference on Bioinformatics and
  Biomedicine (BIBM)}, pages 374--379. IEEE, 2022.

\bibitem{fan2021propagation}
Xiaolong Fan, Maoguo Gong, Yue Wu, AK~Qin, and Yu~Xie.
\newblock Propagation enhanced neural message passing for graph representation
  learning.
\newblock {\em IEEE Transactions on Knowledge and Data Engineering}, 2021.

\bibitem{li2021introducing}
Yuquan Li, Pengyong Li, Xing Yang, Chang-Yu Hsieh, Shengyu Zhang, Xiaorui Wang,
  Ruiqiang Lu, Huanxiang Liu, and Xiaojun Yao.
\newblock Introducing block design in graph neural networks for molecular
  properties prediction.
\newblock {\em Chemical Engineering Journal}, 414:128817, 2021.

\bibitem{ma2020multi}
Hehuan Ma, Yatao Bian, Yu~Rong, Wenbing Huang, Tingyang Xu, Weiyang Xie, Geyan
  Ye, and Junzhou Huang.
\newblock Multi-view graph neural networks for molecular property prediction.
\newblock {\em arXiv preprint arXiv:2005.13607}, 2020.

\bibitem{liu2021hypergraph}
Xiang Liu, Xiangjun Wang, Jie Wu, and Kelin Xia.
\newblock Hypergraph-based persistent cohomology (hpc) for molecular
  representations in drug design.
\newblock {\em Briefings in Bioinformatics}, 22(5):bbaa411, 2021.

\bibitem{feng2022mgmae}
Jinjia Feng, Zhen Wang, Yaliang Li, Bolin Ding, Zhewei Wei, and Hongteng Xu.
\newblock Mgmae: Molecular representation learning by reconstructing
  heterogeneous graphs with a high mask ratio.
\newblock In {\em Proceedings of the 31st ACM International Conference on
  Information \& Knowledge Management}, pages 509--519, 2022.

\bibitem{hasebe2021knowledge}
Tatsuya Hasebe.
\newblock Knowledge-embedded message-passing neural networks: Improving
  molecular property prediction with human knowledge.
\newblock {\em ACS omega}, 6(42):27955--27967, 2021.

\bibitem{yang2021deep}
Shuwen Yang, Ziyao Li, Guojie Song, and Lingsheng Cai.
\newblock Deep molecular representation learning via fusing physical and
  chemical information.
\newblock {\em Advances in Neural Information Processing Systems},
  34:16346--16357, 2021.

\bibitem{zang2023hierarchical}
Xuan Zang, Xianbing Zhao, and Buzhou Tang.
\newblock Hierarchical molecular graph self-supervised learning for property
  prediction.
\newblock {\em Communications Chemistry}, 6(1):34, 2023.

\bibitem{liu2021hierarchical}
Ning Liu, Songlei Jian, Dongsheng Li, Yiming Zhang, Zhiquan Lai, and Hongzuo
  Xu.
\newblock Hierarchical adaptive pooling by capturing high-order dependency for
  graph representation learning.
\newblock {\em IEEE Transactions on Knowledge and Data Engineering}, 2021.

\bibitem{gao2021higher}
Jianliang Gao, Jun Gao, Xiaoting Ying, Mingming Lu, and Jianxin Wang.
\newblock Higher-order interaction goes neural: A substructure assembling graph
  attention network for graph classification.
\newblock {\em IEEE Transactions on Knowledge and Data Engineering}, 2021.

\bibitem{ye2022molecular}
Xian-bin Ye, Quanlong Guan, Weiqi Luo, Liangda Fang, Zhao-Rong Lai, and Jun
  Wang.
\newblock Molecular substructure graph attention network for molecular property
  identification in drug discovery.
\newblock {\em Pattern Recognition}, 128:108659, 2022.

\bibitem{zhu2022hignn}
Weimin Zhu, Yi~Zhang, Duancheng Zhao, Jianrong Xu, and Ling Wang.
\newblock Hignn: A hierarchical informative graph neural network for molecular
  property prediction equipped with feature-wise attention.
\newblock {\em Journal of Chemical Information and Modeling}, 63(1):43--55,
  2022.

\bibitem{lu2019molecular}
Chengqiang Lu, Qi~Liu, Chao Wang, Zhenya Huang, Peize Lin, and Lixin He.
\newblock Molecular property prediction: A multilevel quantum interactions
  modeling perspective.
\newblock In {\em Proceedings of the AAAI conference on artificial
  intelligence}, volume~33, pages 1052--1060, 2019.

\bibitem{fey2020hierarchical}
Matthias Fey, Jan-Gin Yuen, and Frank Weichert.
\newblock Hierarchical inter-message passing for learning on molecular graphs.
\newblock {\em arXiv preprint arXiv:2006.12179}, 2020.

\bibitem{wu2023molformer}
Fang Wu, Dragomir Radev, and Stan~Z Li.
\newblock Molformer: Motif-based transformer on 3d heterogeneous molecular
  graphs.
\newblock In {\em Proceedings of the AAAI Conference on Artificial
  Intelligence}, volume~37, pages 5312--5320, 2023.

\bibitem{fuchs2020se}
Fabian Fuchs, Daniel Worrall, Volker Fischer, and Max Welling.
\newblock Se (3)-transformers: 3d roto-translation equivariant attention
  networks.
\newblock {\em Advances in neural information processing systems},
  33:1970--1981, 2020.

\bibitem{schutt2021equivariant}
Kristof Sch{\"u}tt, Oliver Unke, and Michael Gastegger.
\newblock Equivariant message passing for the prediction of tensorial
  properties and molecular spectra.
\newblock In {\em International Conference on Machine Learning}, pages
  9377--9388. PMLR, 2021.

\bibitem{brandstetter2021geometric}
Johannes Brandstetter, Rob Hesselink, Elise van~der Pol, Erik~J Bekkers, and
  Max Welling.
\newblock Geometric and physical quantities improve e (3) equivariant message
  passing.
\newblock {\em arXiv preprint arXiv:2110.02905}, 2021.

\bibitem{gasteiger2021gemnet}
Johannes Gasteiger, Florian Becker, and Stephan G{\"u}nnemann.
\newblock Gemnet: Universal directional graph neural networks for molecules.
\newblock {\em Advances in Neural Information Processing Systems},
  34:6790--6802, 2021.

\bibitem{gasteiger2020fast}
Johannes Gasteiger, Shankari Giri, Johannes~T Margraf, and Stephan
  G{\"u}nnemann.
\newblock Fast and uncertainty-aware directional message passing for
  non-equilibrium molecules.
\newblock {\em arXiv preprint arXiv:2011.14115}, 2020.

\bibitem{shuaibi2021rotation}
Muhammed Shuaibi, Adeesh Kolluru, Abhishek Das, Aditya Grover, Anuroop Sriram,
  Zachary Ulissi, and C~Lawrence Zitnick.
\newblock Rotation invariant graph neural networks using spin convolutions.
\newblock {\em arXiv preprint arXiv:2106.09575}, 2021.

\bibitem{wang2019smiles}
Sheng Wang, Yuzhi Guo, Yuhong Wang, Hongmao Sun, and Junzhou Huang.
\newblock Smiles-bert: large scale unsupervised pre-training for molecular
  property prediction.
\newblock In {\em Proceedings of the 10th ACM international conference on
  bioinformatics, computational biology and health informatics}, pages
  429--436, 2019.

\bibitem{wang2021molcloze}
Yingheng Wang, Xin Chen, Yaosen Min, and Ji~Wu.
\newblock Molcloze: a unified cloze-style self-supervised molecular structure
  learning model for chemical property prediction.
\newblock In {\em 2021 IEEE International Conference on Bioinformatics and
  Biomedicine (BIBM)}, pages 2896--2903. IEEE, 2021.

\bibitem{winter2022smile}
Benedikt Winter, Clemens Winter, Johannes Schilling, and Andr{\'e} Bardow.
\newblock A smile is all you need: predicting limiting activity coefficients
  from smiles with natural language processing.
\newblock {\em Digital Discovery}, 1(6):859--869, 2022.

\bibitem{su2024roformer}
Jianlin Su, Murtadha Ahmed, Yu~Lu, Shengfeng Pan, Wen Bo, and Yunfeng Liu.
\newblock Roformer: Enhanced transformer with rotary position embedding.
\newblock {\em Neurocomputing}, 568:127063, 2024.

\bibitem{maziarka2020molecule}
{\L}ukasz Maziarka, Tomasz Danel, S{\l}awomir Mucha, Krzysztof Rataj, Jacek
  Tabor, and Stanis{\l}aw Jastrz{{e}}bski.
\newblock Molecule attention transformer.
\newblock {\em arXiv preprint arXiv:2002.08264}, 2020.

\bibitem{park2022grpe}
Wonpyo Park, Woonggi Chang, Donggeon Lee, Juntae Kim, and Seung-won Hwang.
\newblock Grpe: Relative positional encoding for graph transformer.
\newblock {\em arXiv preprint arXiv:2201.12787}, 2022.

\bibitem{hussain2022global}
Md~Shamim Hussain, Mohammed~J Zaki, and Dharmashankar Subramanian.
\newblock Global self-attention as a replacement for graph convolution.
\newblock In {\em Proceedings of the 28th ACM SIGKDD Conference on Knowledge
  Discovery and Data Mining}, pages 655--665, 2022.

\bibitem{masters2022gps++}
Dominic Masters, Josef Dean, Kerstin Klaser, Zhiyi Li, Sam Maddrell-Mander,
  Adam Sanders, Hatem Helal, Deniz Beker, Ladislav Ramp{\'a}{\v{s}}ek, and
  Dominique Beaini.
\newblock Gps++: An optimised hybrid mpnn/transformer for molecular property
  prediction.
\newblock {\em arXiv preprint arXiv:2212.02229}, 2022.

\bibitem{chen2023graph}
Zhe Chen, Hao Tan, Tao Wang, Tianrun Shen, Tong Lu, Qiuying Peng, Cheng Cheng,
  and Yue Qi.
\newblock Graph propagation transformer for graph representation learning.
\newblock {\em arXiv preprint arXiv:2305.11424}, 2023.

\bibitem{ren2023enhancing}
Gao-Peng Ren, Ke-Jun Wu, and Yuchen He.
\newblock Enhancing molecular representations via graph transformation layers.
\newblock {\em Journal of Chemical Information and Modeling}, 63(9):2679--2688,
  2023.

\bibitem{gao2023transfoxmol}
Jian Gao, Zheyuan Shen, Yufeng Xie, Jialiang Lu, Yang Lu, Sikang Chen, Qingyu
  Bian, Yue Guo, Liteng Shen, Jian Wu, et~al.
\newblock Transfoxmol: predicting molecular property with focused attention.
\newblock {\em Briefings in Bioinformatics}, 24(5):bbad306, 2023.

\bibitem{jiang2023pharmacophoric}
Yinghui Jiang, Shuting Jin, Xurui Jin, Xianglu Xiao, Wenfan Wu, Xiangrong Liu,
  Qiang Zhang, Xiangxiang Zeng, Guang Yang, and Zhangming Niu.
\newblock Pharmacophoric-constrained heterogeneous graph transformer model for
  molecular property prediction.
\newblock {\em Communications Chemistry}, 6(1):60, 2023.

\bibitem{hirohara2018convolutional}
Maya Hirohara, Yutaka Saito, Yuki Koda, Kengo Sato, and Yasubumi Sakakibara.
\newblock Convolutional neural network based on smiles representation of
  compounds for detecting chemical motif.
\newblock {\em BMC bioinformatics}, 19:83--94, 2018.

\bibitem{jiang2022molecular}
Peiran Jiang, Ying Chi, Xiao-Shuang Li, Zhenyu Meng, Xiang Liu, Xian-Sheng Hua,
  and Kelin Xia.
\newblock Molecular persistent spectral image (mol-psi) representation for
  machine learning models in drug design.
\newblock {\em Briefings in Bioinformatics}, 23(1):bbab527, 2022.

\bibitem{kuzminykh20183d}
Denis Kuzminykh, Daniil Polykovskiy, Artur Kadurin, Alexander Zhebrak, Ivan
  Baskov, Sergey Nikolenko, Rim Shayakhmetov, and Alex Zhavoronkov.
\newblock 3d molecular representations based on the wave transform for
  convolutional neural networks.
\newblock {\em Molecular pharmaceutics}, 15(10):4378--4385, 2018.

\bibitem{cai2022fp}
Hanxuan Cai, Huimin Zhang, Duancheng Zhao, Jingxing Wu, and Ling Wang.
\newblock Fp-gnn: a versatile deep learning architecture for enhanced molecular
  property prediction.
\newblock {\em Briefings in bioinformatics}, 23(6):bbac408, 2022.

\bibitem{wang2019molecule}
Xiaofeng Wang, Zhen Li, Mingjian Jiang, Shuang Wang, Shugang Zhang, and
  Zhiqiang Wei.
\newblock Molecule property prediction based on spatial graph embedding.
\newblock {\em Journal of chemical information and modeling}, 59(9):3817--3828,
  2019.

\bibitem{liu2023prediction}
Jianping Liu, Xiujuan Lei, Yuchen Zhang, and Yi~Pan.
\newblock The prediction of molecular toxicity based on bigru and graphsage.
\newblock {\em Computers in Biology and Medicine}, 153:106524, 2023.

\bibitem{luo2023molfm}
Yizhen Luo, Kai Yang, Massimo Hong, Xingyi Liu, and Zaiqing Nie.
\newblock Molfm: A multimodal molecular foundation model.
\newblock {\em arXiv preprint arXiv:2307.09484}, 2023.

\bibitem{sun2022molecular}
Yan Sun, Mohaiminul Islam, Ehsan Zahedi, M{\'e}laine Kuenemann, Hassan Chouaib,
  and Pingzhao Hu.
\newblock Molecular property prediction based on bimodal supervised contrastive
  learning.
\newblock In {\em 2022 IEEE International Conference on Bioinformatics and
  Biomedicine (BIBM)}, pages 394--397. IEEE, 2022.

\bibitem{liu2024git}
Pengfei Liu, Yiming Ren, Jun Tao, and Zhixiang Ren.
\newblock Git-mol: A multi-modal large language model for molecular science
  with graph, image, and text.
\newblock {\em Computers in Biology and Medicine}, 171:108073, 2024.

\bibitem{tang2022merged}
Qiang Tang, Fulei Nie, Qi~Zhao, and Wei Chen.
\newblock A merged molecular representation deep learning method for
  blood--brain barrier permeability prediction.
\newblock {\em Briefings in Bioinformatics}, 23(5):bbac357, 2022.

\bibitem{zhang2023transg}
Taohong Zhang, Saian Chen, Aziguli Wulamu, Xuxu Guo, Qianqian Li, and Han
  Zheng.
\newblock Transg-net: transformer and graph neural network based multi-modal
  data fusion network for molecular properties prediction.
\newblock {\em Applied Intelligence}, 53(12):16077--16088, 2023.

\bibitem{chen2021algebraic}
Dong Chen, Kaifu Gao, Duc~Duy Nguyen, Xin Chen, Yi~Jiang, Guo-Wei Wei, and Feng
  Pan.
\newblock Algebraic graph-assisted bidirectional transformers for molecular
  property prediction.
\newblock {\em Nature communications}, 12(1):3521, 2021.

\bibitem{shen2021out}
Wan~Xiang Shen, Xian Zeng, Feng Zhu, Ya~Li Wang, Chu Qin, Ying Tan, Yu~Yang
  Jiang, and Yu~Zong Chen.
\newblock Out-of-the-box deep learning prediction of pharmaceutical properties
  by broadly learned knowledge-based molecular representations.
\newblock {\em Nature Machine Intelligence}, 3(4):334--343, 2021.

\bibitem{liu2022spherical}
Yi~Liu, Limei Wang, Meng Liu, Yuchao Lin, Xuan Zhang, Bora Oztekin, and
  Shuiwang Ji.
\newblock Spherical message passing for 3d molecular graphs.
\newblock In {\em International Conference on Learning Representations (ICLR)},
  2022.

\bibitem{wang2022advanced}
Zhengyang Wang, Meng Liu, Youzhi Luo, Zhao Xu, Yaochen Xie, Limei Wang, Lei
  Cai, Qi~Qi, Zhuoning Yuan, Tianbao Yang, et~al.
\newblock Advanced graph and sequence neural networks for molecular property
  prediction and drug discovery.
\newblock {\em Bioinformatics}, 38(9):2579--2586, 2022.

\bibitem{zhu2023dual}
Jinhua Zhu, Yingce Xia, Lijun Wu, Shufang Xie, Wengang Zhou, Tao Qin, Houqiang
  Li, and Tie-Yan Liu.
\newblock Dual-view molecular pre-training.
\newblock In {\em Proceedings of the 29th ACM SIGKDD Conference on Knowledge
  Discovery and Data Mining}, pages 3615--3627, 2023.

\bibitem{sun2021mocl}
Mengying Sun, Jing Xing, Huijun Wang, Bin Chen, and Jiayu Zhou.
\newblock Mocl: data-driven molecular fingerprint via knowledge-aware
  contrastive learning from molecular graph.
\newblock In {\em Proceedings of the 27th ACM SIGKDD Conference on Knowledge
  Discovery \& Data Mining}, pages 3585--3594, 2021.

\bibitem{wang2022improving}
Yuyang Wang, Rishikesh Magar, Chen Liang, and Amir Barati~Farimani.
\newblock Improving molecular contrastive learning via faulty negative
  mitigation and decomposed fragment contrast.
\newblock {\em Journal of Chemical Information and Modeling},
  62(11):2713--2725, 2022.

\bibitem{you2020graph}
Yuning You, Tianlong Chen, Yongduo Sui, Ting Chen, Zhangyang Wang, and Yang
  Shen.
\newblock Graph contrastive learning with augmentations.
\newblock {\em Advances in neural information processing systems},
  33:5812--5823, 2020.

\bibitem{xia2022mole}
Jun Xia, Chengshuai Zhao, Bozhen Hu, Zhangyang Gao, Cheng Tan, Yue Liu, Siyuan
  Li, and Stan~Z Li.
\newblock Mole-bert: Rethinking pre-training graph neural networks for
  molecules.
\newblock In {\em The Eleventh International Conference on Learning
  Representations}, 2022.

\bibitem{wu2022knowledge}
Zhenxing Wu, Dejun Jiang, Jike Wang, Xujun Zhang, Hongyan Du, Lurong Pan,
  Chang-Yu Hsieh, Dongsheng Cao, and Tingjun Hou.
\newblock Knowledge-based bert: a method to extract molecular features like
  computational chemists.
\newblock {\em Briefings in Bioinformatics}, 23(3):bbac131, 2022.

\bibitem{wu2023chemistry}
Zhenxing Wu, Jike Wang, Hongyan Du, Dejun Jiang, Yu~Kang, Dan Li, Peichen Pan,
  Yafeng Deng, Dongsheng Cao, Chang-Yu Hsieh, et~al.
\newblock Chemistry-intuitive explanation of graph neural networks for
  molecular property prediction with substructure masking.
\newblock {\em Nature Communications}, 14(1):2585, 2023.

\bibitem{kim2023fragment}
Seojin Kim, Jaehyun Nam, Junsu Kim, Hankook Lee, Sungsoo Ahn, and Jinwoo Shin.
\newblock Fragment-based multi-view molecular contrastive learning.
\newblock In {\em Workshop on''Machine Learning for Materials''ICLR 2023},
  2023.

\bibitem{wu2023instructbio}
Fang Wu, Huiling Qin, Wenhao Gao, Siyuan Li, Connor~W Coley, Stan~Z Li,
  Xianyuan Zhan, and Jinbo Xu.
\newblock Instructbio: A large-scale semi-supervised learning paradigm for
  biochemical problems.
\newblock {\em arXiv preprint arXiv:2304.03906}, 2023.

\bibitem{lv2023meta}
Qiujie Lv, Guanxing Chen, Ziduo Yang, Weihe Zhong, and Calvin Yu-Chian Chen.
\newblock Meta learning with graph attention networks for low-data drug
  discovery.
\newblock {\em IEEE Transactions on Neural Networks and Learning Systems},
  2023.

\bibitem{devlinBERTPretrainingDeep2018}
Jacob Devlin, Ming-Wei Chang, Kenton Lee, and Kristina Toutanova.
\newblock {{BERT}}: {{Pre-training}} of {{Deep Bidirectional Transformers}} for
  {{Language Understanding}}.
\newblock 2018.

\bibitem{floridi2020gpt}
Luciano Floridi and Massimo Chiriatti.
\newblock Gpt-3: Its nature, scope, limits, and consequences.
\newblock {\em Minds and Machines}, 30:681--694, 2020.

\bibitem{zhang2021mg}
Xiao-Chen Zhang, Cheng-Kun Wu, Zhi-Jiang Yang, Zhen-Xing Wu, Jia-Cai Yi,
  Chang-Yu Hsieh, Ting-Jun Hou, and Dong-Sheng Cao.
\newblock Mg-bert: leveraging unsupervised atomic representation learning for
  molecular property prediction.
\newblock {\em Briefings in bioinformatics}, 22(6):bbab152, 2021.

\bibitem{ahmad2022chemberta}
Walid Ahmad, Elana Simon, Seyone Chithrananda, Gabriel Grand, and Bharath
  Ramsundar.
\newblock Chemberta-2: Towards chemical foundation models.
\newblock {\em arXiv preprint arXiv:2209.01712}, 2022.

\bibitem{irwin2022chemformer}
Ross Irwin, Spyridon Dimitriadis, Jiazhen He, and Esben~Jannik Bjerrum.
\newblock Chemformer: a pre-trained transformer for computational chemistry.
\newblock {\em Machine Learning: Science and Technology}, 3(1):015022, 2022.

\bibitem{hu2019strategies}
Weihua Hu, Bowen Liu, Joseph Gomes, Marinka Zitnik, Percy Liang, Vijay Pande,
  and Jure Leskovec.
\newblock Strategies for pre-training graph neural networks.
\newblock {\em arXiv preprint arXiv:1905.12265}, 2019.

\bibitem{godwin2021simple}
Jonathan Godwin, Michael Schaarschmidt, Alexander Gaunt, Alvaro
  Sanchez-Gonzalez, Yulia Rubanova, Petar Veli{\v{c}}kovi{\'c}, James
  Kirkpatrick, and Peter Battaglia.
\newblock Simple gnn regularisation for 3d molecular property prediction \&
  beyond.
\newblock {\em arXiv preprint arXiv:2106.07971}, 2021.

\bibitem{liu2022molecular}
Shengchao Liu, Hongyu Guo, and Jian Tang.
\newblock Molecular geometry pretraining with se (3)-invariant denoising
  distance matching.
\newblock {\em arXiv preprint arXiv:2206.13602}, 2022.

\bibitem{feng2023fractional}
Shikun Feng, Yuyan Ni, Yanyan Lan, Zhi-Ming Ma, and Wei-Ying Ma.
\newblock Fractional denoising for 3d molecular pre-training.
\newblock In {\em International Conference on Machine Learning}, pages
  9938--9961. PMLR, 2023.

\bibitem{jiao2023energy}
Rui Jiao, Jiaqi Han, Wenbing Huang, Yu~Rong, and Yang Liu.
\newblock Energy-motivated equivariant pretraining for 3d molecular graphs.
\newblock In {\em Proceedings of the AAAI Conference on Artificial
  Intelligence}, volume~37, pages 8096--8104, 2023.

\bibitem{gao2022supervised}
Xiang Gao, Weihao Gao, Wenzhi Xiao, Zhirui Wang, Chong Wang, and Liang Xiang.
\newblock Supervised pretraining for molecular force fields and properties
  prediction.
\newblock {\em arXiv preprint arXiv:2211.14429}, 2022.

\bibitem{wang2023automated}
Xu~Wang, Huan Zhao, Wei-wei Tu, and Quanming Yao.
\newblock Automated 3d pre-training for molecular property prediction.
\newblock In {\em Proceedings of the 29th ACM SIGKDD Conference on Knowledge
  Discovery and Data Mining}, pages 2419--2430, 2023.

\bibitem{zeng2023molkd}
Liang Zeng, Lanqing Li, and Jian Li.
\newblock Molkd: Distilling cross-modal knowledge in chemical reactions for
  molecular property prediction.
\newblock {\em arXiv preprint arXiv:2305.01912}, 2023.

\bibitem{broberg2022pre}
Johan Broberg, Maria B{\aa}nkestad, and Erik Ylip{\"a}{\"a}.
\newblock Pre-training transformers for molecular property prediction using
  reaction prediction.
\newblock {\em arXiv preprint arXiv:2207.02724}, 2022.

\bibitem{zhang2022pushing}
Xiao-Chen Zhang, Cheng-Kun Wu, Jia-Cai Yi, Xiang-Xiang Zeng, Can-Qun Yang,
  Ai-Ping Lu, Ting-Jun Hou, and Dong-Sheng Cao.
\newblock Pushing the boundaries of molecular property prediction for drug
  discovery with multitask learning bert enhanced by smiles enumeration.
\newblock {\em Research}, 2022:0004, 2022.

\bibitem{abdel2022large}
Hisham Abdel-Aty and Ian~R Gould.
\newblock Large-scale distributed training of transformers for chemical
  fingerprinting.
\newblock {\em Journal of Chemical Information and Modeling},
  62(20):4852--4862, 2022.

\bibitem{zheng2023casangcl}
Zixi Zheng, Yanyan Tan, Hong Wang, Shengpeng Yu, Tianyu Liu, and Cheng Liang.
\newblock Casangcl: pre-training and fine-tuning model based on cascaded
  attention network and graph contrastive learning for molecular property
  prediction.
\newblock {\em Briefings in Bioinformatics}, 24(1):bbac566, 2023.

\bibitem{guan2023t}
Xiaoyu Guan and Daoqiang Zhang.
\newblock T-mgcl: Molecule graph contrastive learning based on transformer for
  molecular property prediction.
\newblock {\em IEEE/ACM Transactions on Computational Biology and
  Bioinformatics}, 2023.

\bibitem{liu2022attention}
Hui Liu, Yibiao Huang, Xuejun Liu, and Lei Deng.
\newblock Attention-wise masked graph contrastive learning for predicting
  molecular property.
\newblock {\em Briefings in bioinformatics}, 23(5):bbac303, 2022.

\bibitem{lin2022prototypical}
Shuai Lin, Chen Liu, Pan Zhou, Zi-Yuan Hu, Shuojia Wang, Ruihui Zhao, Yefeng
  Zheng, Liang Lin, Eric Xing, and Xiaodan Liang.
\newblock Prototypical graph contrastive learning.
\newblock {\em IEEE Transactions on Neural Networks and Learning Systems},
  2022.

\bibitem{cui2023mocgcl}
Jinhao Cui, Heyan Chai, Yanbin Gong, Ye~Ding, Zhongyun Hua, Cuiyun Gao, and
  Qing Liao.
\newblock Mocgcl: Molecular graph contrastive learning via negative selection.
\newblock In {\em 2023 International Joint Conference on Neural Networks
  (IJCNN)}, pages 1--8. IEEE, 2023.

\bibitem{he2020momentum}
Kaiming He, Haoqi Fan, Yuxin Wu, Saining Xie, and Ross Girshick.
\newblock Momentum contrast for unsupervised visual representation learning.
\newblock In {\em Proceedings of the IEEE/CVF conference on computer vision and
  pattern recognition}, pages 9729--9738, 2020.

\bibitem{zaki2014data}
Mohammed~J Zaki and Wagner Meira.
\newblock {\em Data mining and analysis: fundamental concepts and algorithms}.
\newblock Cambridge University Press, 2014.

\bibitem{wang2021molecular}
Yingheng Wang, Yaosen Min, Erzhuo Shao, and Ji~Wu.
\newblock Molecular graph contrastive learning with parameterized explainable
  augmentations.
\newblock In {\em 2021 IEEE International Conference on Bioinformatics and
  Biomedicine (BIBM)}, pages 1558--1563. IEEE, 2021.

\bibitem{liu2022hiermrl}
Maotao Liu, Yifan Yang, Xu~Gong, Li~Liu, and Qun Liu.
\newblock Hiermrl: Hierarchical structure-aware molecular representation
  learning for property prediction.
\newblock In {\em 2022 IEEE International Conference on Bioinformatics and
  Biomedicine (BIBM)}, pages 386--389. IEEE, 2022.

\bibitem{wang2023molecular}
Jinxian Wang, Jihong Guan, and Shuigeng Zhou.
\newblock Molecular property prediction by contrastive learning with
  attention-guided positive sample selection.
\newblock {\em Bioinformatics}, 39(5):btad258, 2023.

\bibitem{moon20233d}
Kisung Moon, Hyeon-Jin Im, and Sunyoung Kwon.
\newblock 3d graph contrastive learning for molecular property prediction.
\newblock {\em Bioinformatics}, 39(6):btad371, 2023.

\bibitem{kuang20233d}
Taojie Kuang, Yiming Ren, and Zhixiang Ren.
\newblock 3d-mol: A novel contrastive learning framework for molecular property
  prediction with 3d information.
\newblock {\em arXiv preprint arXiv:2309.17366}, 2023.

\bibitem{wu2023medical}
Xuehong Wu, Junwen Duan, Yi~Pan, and Min Li.
\newblock Medical knowledge graph: Data sources, construction, reasoning, and
  applications.
\newblock {\em Big Data Mining and Analytics}, 6(2):201--217, 2023.

\bibitem{hua2022chemical}
Rui Hua, Xinyan Wang, Chuang Cheng, Qiang Zhu, and Xuezhong Zhou.
\newblock A chemical domain knowledge-aware framework for multi-view molecular
  property prediction.
\newblock In {\em China Conference on Knowledge Graph and Semantic Computing},
  pages 1--11. Springer, 2022.

\bibitem{fang2022molecular}
Yin Fang, Qiang Zhang, Haihong Yang, Xiang Zhuang, Shumin Deng, Wen Zhang, Ming
  Qin, Zhuo Chen, Xiaohui Fan, and Huajun Chen.
\newblock Molecular contrastive learning with chemical element knowledge graph.
\newblock In {\em Proceedings of the AAAI Conference on Artificial
  Intelligence}, volume~36, pages 3968--3976, 2022.

\bibitem{xu2021self}
Minghao Xu, Hang Wang, Bingbing Ni, Hongyu Guo, and Jian Tang.
\newblock Self-supervised graph-level representation learning with local and
  global structure.
\newblock In {\em International Conference on Machine Learning}, pages
  11548--11558. PMLR, 2021.

\bibitem{shen2020molgnn}
Xiaoke Shen, Yang Liu, You Wu, and Lei Xie.
\newblock Molgnn: Self-supervised motif learning graph neural network for drug
  discovery.
\newblock In {\em Machine Learning for Molecules Workshop at NeurIPS}, volume
  2020, page~4, 2020.

\bibitem{luo2022clear}
Xiao Luo, Wei Ju, Meng Qu, Yiyang Gu, Chong Chen, Minghua Deng, Xian-Sheng Hua,
  and Ming Zhang.
\newblock Clear: Cluster-enhanced contrast for self-supervised graph
  representation learning.
\newblock {\em IEEE Transactions on Neural Networks and Learning Systems},
  2022.

\bibitem{benjamin2022graph}
Roy Benjamin, Uriel Singer, and Kira Radinsky.
\newblock Graph neural networks pretraining through inherent supervision for
  molecular property prediction.
\newblock In {\em Proceedings of the 31st ACM International Conference on
  Information \& Knowledge Management}, pages 2903--2912, 2022.

\bibitem{shi2023hegcl}
Gen Shi, Yifan Zhu, Jian~K Liu, and Xuesong Li.
\newblock Hegcl: Advance self-supervised learning in heterogeneous graph-level
  representation.
\newblock {\em IEEE Transactions on Neural Networks and Learning Systems},
  2023.

\bibitem{xie2023self}
Ailin Xie, Ziqiao Zhang, Jihong Guan, and Shuigeng Zhou.
\newblock Self-supervised learning with chemistry-aware fragmentation for
  effective molecular property prediction.
\newblock {\em Briefings in Bioinformatics}, 24(5):bbad296, 2023.

\bibitem{ji2022relmole}
Zewei Ji, Runhan Shi, Jiarui Lu, Fang Li, and Yang Yang.
\newblock Relmole: Molecular representation learning based on two-level graph
  similarities.
\newblock {\em Journal of Chemical Information and Modeling},
  62(22):5361--5372, 2022.

\bibitem{hadsell2006dimensionality}
Raia Hadsell, Sumit Chopra, and Yann LeCun.
\newblock Dimensionality reduction by learning an invariant mapping.
\newblock In {\em 2006 IEEE computer society conference on computer vision and
  pattern recognition (CVPR'06)}, volume~2, pages 1735--1742. IEEE, 2006.

\bibitem{pinheiro2022smiclr}
Gabriel~A Pinheiro, Juarez~LF Da~Silva, and Marcos~G Quiles.
\newblock Smiclr: Contrastive learning on multiple molecular representations
  for semisupervised and unsupervised representation learning.
\newblock {\em Journal of Chemical Information and Modeling},
  62(17):3948--3960, 2022.

\bibitem{zhang2022pseudo}
Chaoran Zhang, Xiangfeng Yan, and Yong Liu.
\newblock Pseudo-siamese neural network based graph and sequence representation
  learning for molecular property prediction.
\newblock In {\em 2022 IEEE International Conference on Bioinformatics and
  Biomedicine (BIBM)}, pages 3911--3913. IEEE, 2022.

\bibitem{stark20223d}
Hannes St{\"a}rk, Dominique Beaini, Gabriele Corso, Prudencio Tossou, Christian
  Dallago, Stephan G{\"u}nnemann, and Pietro Li{\`o}.
\newblock 3d infomax improves gnns for molecular property prediction.
\newblock In {\em International Conference on Machine Learning}, pages
  20479--20502. PMLR, 2022.

\bibitem{zhu2024molecular}
Yanqiao Zhu, Dingshuo Chen, Yuanqi Du, Yingze Wang, Qiang Liu, and Shu Wu.
\newblock Molecular contrastive pretraining with collaborative featurizations.
\newblock {\em Journal of Chemical Information and Modeling}, 64(4):1112--1122,
  2024.
\newblock PMID: 38315002.

\bibitem{tarvainen2017mean}
Antti Tarvainen and Harri Valpola.
\newblock Mean teachers are better role models: Weight-averaged consistency
  targets improve semi-supervised deep learning results.
\newblock {\em Advances in neural information processing systems}, 30, 2017.

\bibitem{chen2021chemical}
Jiarui Chen, Yain-Whar Si, Chon-Wai Un, and Shirley~WI Siu.
\newblock Chemical toxicity prediction based on semi-supervised learning and
  graph convolutional neural network.
\newblock {\em Journal of cheminformatics}, 13(1):1--16, 2021.

\bibitem{berthelot2019mixmatch}
David Berthelot, Nicholas Carlini, Ian Goodfellow, Nicolas Papernot, Avital
  Oliver, and Colin~A Raffel.
\newblock Mixmatch: A holistic approach to semi-supervised learning.
\newblock {\em Advances in neural information processing systems}, 32, 2019.

\bibitem{yu2020semi}
Ke~Yu, Shyam Visweswaran, and Kayhan Batmanghelich.
\newblock Semi-supervised hierarchical drug embedding in hyperbolic space.
\newblock {\em Journal of chemical information and modeling},
  60(12):5647--5657, 2020.

\bibitem{ma2022robust}
Hehuan Ma, Feng Jiang, Yu~Rong, Yuzhi Guo, and Junzhou Huang.
\newblock Robust self-training strategy for various molecular biology
  prediction tasks.
\newblock In {\em Proceedings of the 13th ACM International Conference on
  Bioinformatics, Computational Biology and Health Informatics}, pages 1--5,
  2022.

\bibitem{zhang2018generalized}
Zhilu Zhang and Mert Sabuncu.
\newblock Generalized cross entropy loss for training deep neural networks with
  noisy labels.
\newblock {\em Advances in neural information processing systems}, 31, 2018.

\bibitem{liu2023semi}
Gang Liu, Tong Zhao, Eric Inae, Tengfei Luo, and Meng Jiang.
\newblock Semi-supervised graph imbalanced regression.
\newblock {\em arXiv preprint arXiv:2305.12087}, 2023.

\bibitem{zamir2018taskonomy}
Amir~R Zamir, Alexander Sax, William Shen, Leonidas~J Guibas, Jitendra Malik,
  and Silvio Savarese.
\newblock Taskonomy: Disentangling task transfer learning.
\newblock In {\em Proceedings of the IEEE conference on computer vision and
  pattern recognition}, pages 3712--3722, 2018.

\bibitem{li2019deepchemstable}
Xiuming Li, Xin Yan, Qiong Gu, Huihao Zhou, Di~Wu, and Jun Xu.
\newblock Deepchemstable: chemical stability prediction with an attention-based
  graph convolution network.
\newblock {\em Journal of chemical information and modeling}, 59(3):1044--1049,
  2019.

\bibitem{chen2021exploring}
Xinlei Chen and Kaiming He.
\newblock Exploring simple siamese representation learning.
\newblock In {\em Proceedings of the IEEE/CVF conference on computer vision and
  pattern recognition}, pages 15750--15758, 2021.

\bibitem{li2022improving}
Han Li, Xinyi Zhao, Shuya Li, Fangping Wan, Dan Zhao, and Jianyang Zeng.
\newblock Improving molecular property prediction through a task similarity
  enhanced transfer learning strategy.
\newblock {\em Iscience}, 25(10), 2022.

\bibitem{ju2023few}
Wei Ju, Zequn Liu, Yifang Qin, Bin Feng, Chen Wang, Zhihui Guo, Xiao Luo, and
  Ming Zhang.
\newblock Few-shot molecular property prediction via hierarchically structured
  learning on relation graphs.
\newblock {\em Neural Networks}, 163:122--131, 2023.

\bibitem{nguyen2020meta}
Cuong~Q Nguyen, Constantine Kreatsoulas, and Kim~M Branson.
\newblock Meta-learning gnn initializations for low-resource molecular property
  prediction.
\newblock {\em arXiv preprint arXiv:2003.05996}, 2020.

\bibitem{torres2023few}
Luis Torres, Joel~P Arrais, and Bernardete Ribeiro.
\newblock Few-shot learning via graph embeddings with convolutional networks
  for low-data molecular property prediction.
\newblock {\em Neural Computing and Applications}, 35(18):13167--13185, 2023.

\bibitem{de2022graph}
Haitz~S{\'a}ez de~Oc{\'a}riz~Borde and Federico Barbero.
\newblock Graph neural network expressivity and meta-learning for molecular
  property regression.
\newblock In {\em The First Learning on Graphs Conference}, 2022.

\bibitem{ham2023evidential}
Kyung~Pyo Ham and Lee Sael.
\newblock Evidential meta-model for molecular property prediction.
\newblock {\em Bioinformatics}, 39(10):btad604, 2023.

\bibitem{meng2023meta}
Ziqiao Meng, Yaoman Li, Peilin Zhao, Yang Yu, and Irwin King.
\newblock Meta-learning with motif-based task augmentation for few-shot
  molecular property prediction.
\newblock In {\em Proceedings of the 2023 SIAM International Conference on Data
  Mining (SDM)}, pages 811--819. SIAM, 2023.

\bibitem{guo2021few}
Zhichun Guo, Chuxu Zhang, Wenhao Yu, John Herr, Olaf Wiest, Meng Jiang, and
  Nitesh~V Chawla.
\newblock Few-shot graph learning for molecular property prediction.
\newblock In {\em Proceedings of the web conference 2021}, pages 2559--2567,
  2021.

\bibitem{wang2021property}
Yaqing Wang, Abulikemu Abuduweili, Quanming Yao, and Dejing Dou.
\newblock Property-aware relation networks for few-shot molecular property
  prediction.
\newblock {\em Advances in Neural Information Processing Systems},
  34:17441--17454, 2021.

\bibitem{yao2022chemical}
Shaolun Yao, Zunlei Feng, Jie Song, Lingxiang Jia, Zipeng Zhong, and Mingli
  Song.
\newblock Chemical property relation guided few-shot molecular property
  prediction.
\newblock In {\em 2022 International Joint Conference on Neural Networks
  (IJCNN)}, pages 1--8. IEEE, 2022.

\bibitem{dong2018admetlab}
Jie Dong, Ning-Ning Wang, Zhi-Jiang Yao, Lin Zhang, Yan Cheng, Defang Ouyang,
  Ai-Ping Lu, and Dong-Sheng Cao.
\newblock Admetlab: a platform for systematic admet evaluation based on a
  comprehensively collected admet database.
\newblock {\em Journal of cheminformatics}, 10:1--11, 2018.

\bibitem{van2022exposing}
Derek van Tilborg, Alisa Alenicheva, and Francesca Grisoni.
\newblock Exposing the limitations of molecular machine learning with activity
  cliffs.
\newblock {\em Journal of Chemical Information and Modeling},
  62(23):5938--5951, 2022.

\bibitem{ji2023drugood}
Yuanfeng Ji, Lu~Zhang, Jiaxiang Wu, Bingzhe Wu, Lanqing Li, Long-Kai Huang,
  Tingyang Xu, Yu~Rong, Jie Ren, Ding Xue, et~al.
\newblock Drugood: Out-of-distribution dataset curator and benchmark for
  ai-aided drug discovery--a focus on affinity prediction problems with noise
  annotations.
\newblock In {\em Proceedings of the AAAI Conference on Artificial
  Intelligence}, volume~37, pages 8023--8031, 2023.

\bibitem{chmiela2017machine}
Stefan Chmiela, Alexandre Tkatchenko, Huziel~E Sauceda, Igor Poltavsky,
  Kristof~T Sch{\"u}tt, and Klaus-Robert M{\"u}ller.
\newblock Machine learning of accurate energy-conserving molecular force
  fields.
\newblock {\em Science advances}, 3(5):e1603015, 2017.

\bibitem{morris2020tudataset}
Christopher Morris, Nils~M Kriege, Franka Bause, Kristian Kersting, Petra
  Mutzel, and Marion Neumann.
\newblock Tudataset: A collection of benchmark datasets for learning with
  graphs.
\newblock {\em arXiv preprint arXiv:2007.08663}, 2020.

\bibitem{hu2020open}
Weihua Hu, Matthias Fey, Marinka Zitnik, Yuxiao Dong, Hongyu Ren, Bowen Liu,
  Michele Catasta, and Jure Leskovec.
\newblock Open graph benchmark: Datasets for machine learning on graphs.
\newblock {\em Advances in neural information processing systems},
  33:22118--22133, 2020.

\bibitem{wojtuch2023extended}
Agnieszka Wojtuch, Tomasz Danel, Sabina Podlewska, and {\L}ukasz Maziarka.
\newblock Extended study on atomic featurization in graph neural networks for
  molecular property prediction.
\newblock {\em Journal of Cheminformatics}, 15(1):81, 2023.

\bibitem{zeng2022deep}
Zheni Zeng, Yuan Yao, Zhiyuan Liu, and Maosong Sun.
\newblock A deep-learning system bridging molecule structure and biomedical
  text with comprehension comparable to human professionals.
\newblock {\em Nature communications}, 13(1):862, 2022.

\bibitem{xu2018powerful}
Keyulu Xu, Weihua Hu, Jure Leskovec, and Stefanie Jegelka.
\newblock How powerful are graph neural networks?
\newblock In {\em International Conference on Learning Representations}, 2018.

\bibitem{su2022molecular}
Bing Su, Dazhao Du, Zhao Yang, Yujie Zhou, Jiangmeng Li, Anyi Rao, Hao Sun,
  Zhiwu Lu, and Ji-Rong Wen.
\newblock A molecular multimodal foundation model associating molecule graphs
  with natural language.
\newblock {\em arXiv preprint arXiv:2209.05481}, 2022.

\bibitem{tang2023mollm}
Xiangru Tang, Andrew Tran, Jeffrey Tan, and Mark~B Gerstein.
\newblock Mollm: A unified language model to integrate biomedical text with 2d
  and 3d molecular representations.
\newblock {\em bioRxiv preprint bioRxiv:2023.11.25.568656}, 2023.

\end{thebibliography}
\end{document}